\documentclass[sigconf]{acmart}

\usepackage{stfloats,microtype,balance,enumitem}
\usepackage{color}
\usepackage{microtype}

\AtBeginDocument{%
  \providecommand\BibTeX{{%
    \normalfont B\kern-0.5em{\scshape i\kern-0.25em b}\kern-0.8em\TeX}}}

\copyrightyear{2023}
\acmYear{2023}
\setcopyright{rightsretained}
\acmConference[HRI '23]{Proceedings of the 2023 ACM/IEEE International Conference on Human-Robot Interaction}{March 13--16, 2023}{Stockholm, Sweden}
\acmBooktitle{Proceedings of the 2023 ACM/IEEE International Conference on Human-Robot Interaction (HRI '23), March 13--16, 2023, Stockholm, Sweden}\acmDOI{10.1145/3568162.3576991}
\acmISBN{978-1-4503-9964-7/23/03}

\acmSubmissionID{1265}

\begin{document}

\newcommand{\tool}{{\textit{Tabula}}}
\newcommand{\act}{{\textit{a}}}
\newcommand{\con}{{\textit{e}}}
\newcommand{\utt}{{$\mu$}}
\newcommand{\skt}{{$\sigma$}}
\newcommand{\sktloop}{{$\sigma'$}}
\newcommand{\rec}{{\textit{R}}}
\newcommand{\trc}{{\textit{t}}}
\newcommand{\Trc}{{\textit{T}}}
\newcommand{\prog}{{\textit{P}}}

\newcommand{\tcmd}[2]{\textbf{\texttt{#1}}\textit{ #2}}

\renewenvironment{quotation}
{\list{}{\leftmargin=12pt
  \listparindent \parindent
  \itemindent \listparindent
  \rightmargin \leftmargin
  \parsep \parskip}%
  \item\relax\noindent\ignorespaces}
{\endlist}

\title{Sketching Robot Programs On the Fly}


\author{David Porfirio}
\orcid{0000-0001-5383-3266}
\affiliation{%
  \institution{Naval Research Laboratory}
  \city{Washington}
  \state{DC}
  \country{United States}
  \postcode{53706}
}
\email{david.porfirio.ctr@nrl.navy.mil}

\author{Laura Stegner}
\orcid{0000-0003-4496-0727}
\affiliation{%
  \institution{University of Wisconsin–Madison}
  \city{Madison}
  \state{Wisconsin}
  \country{United States}
  \postcode{53706}
}
\email{stegner@wisc.edu}

\author{Maya Cakmak}
\orcid{0000-0001-8457-6610}
\affiliation{%
  \institution{University of Washington}
  \city{Seattle}
  \state{Washington}
  \country{United States}
  \postcode{98195}
}
\email{mcakmak@cs.washington.edu}

\author{Allison Sauppé}
\orcid{0000-0002-7548-368X}
\affiliation{%
  \institution{University of Wisconsin–La Crosse}
  \city{La Crosse}
  \state{Wisconsin}
  \country{United States}
  \postcode{54601}
}
\email{asauppe@uwlax.edu}

\author{Aws Albarghouthi}
\orcid{0000-0003-4577-175X}
\affiliation{%
  \institution{University of Wisconsin–Madison}
  \city{Madison}
  \state{Wisconsin}
  \country{United States}
  \postcode{53706}
}
\email{aws@cs.wisc.edu}

\author{Bilge Mutlu}
\orcid{0000-0002-9456-1495}
\affiliation{%
  \institution{University of Wisconsin–Madison}
  \city{Madison}
  \state{Wisconsin}
  \country{United States}
  \postcode{53706}
}
\email{bilge@cs.wisc.edu}

\renewcommand{\shortauthors}{David Porfirio et al.}

\begin{abstract}
Service robots for personal use in the home and the workplace require end-user development solutions for swiftly scripting robot tasks  as the need arises.
Many existing solutions preserve ease, efficiency, and convenience through simple programming interfaces or by restricting task complexity. Others facilitate meticulous task design 
but often do so at the expense of simplicity and efficiency. There is a need for robot programming solutions that reconcile the complexity of robotics with the on-the-fly goals of end-user development.
In  response to this need, we present a novel, multimodal, and on-the-fly development system, \tool{}. Inspired by a formative design study with a prototype, \tool{} leverages a combination of spoken language for specifying the core of a robot task and sketching for contextualizing the core. The result is that developers can script partial, sloppy versions of robot programs to be completed and refined by a program synthesizer. Lastly, we demonstrate our anticipated use cases of \tool{} via a set of application scenarios.
\end{abstract}

\ccsdesc[500]{Human-centered computing~Systems and tools for interaction design}
\ccsdesc[500]{Software and its engineering}

\keywords{human-robot interaction, end-user development, sketching}


\renewcommand\authors{David Porfirio, Laura Stegner, Maya Cakmak, Allison Sauppé, Aws Albarghouthi, and Bilge Mutlu}

\maketitle
\begin{figure}[!t]
    \centering
    \includegraphics[width=\columnwidth]{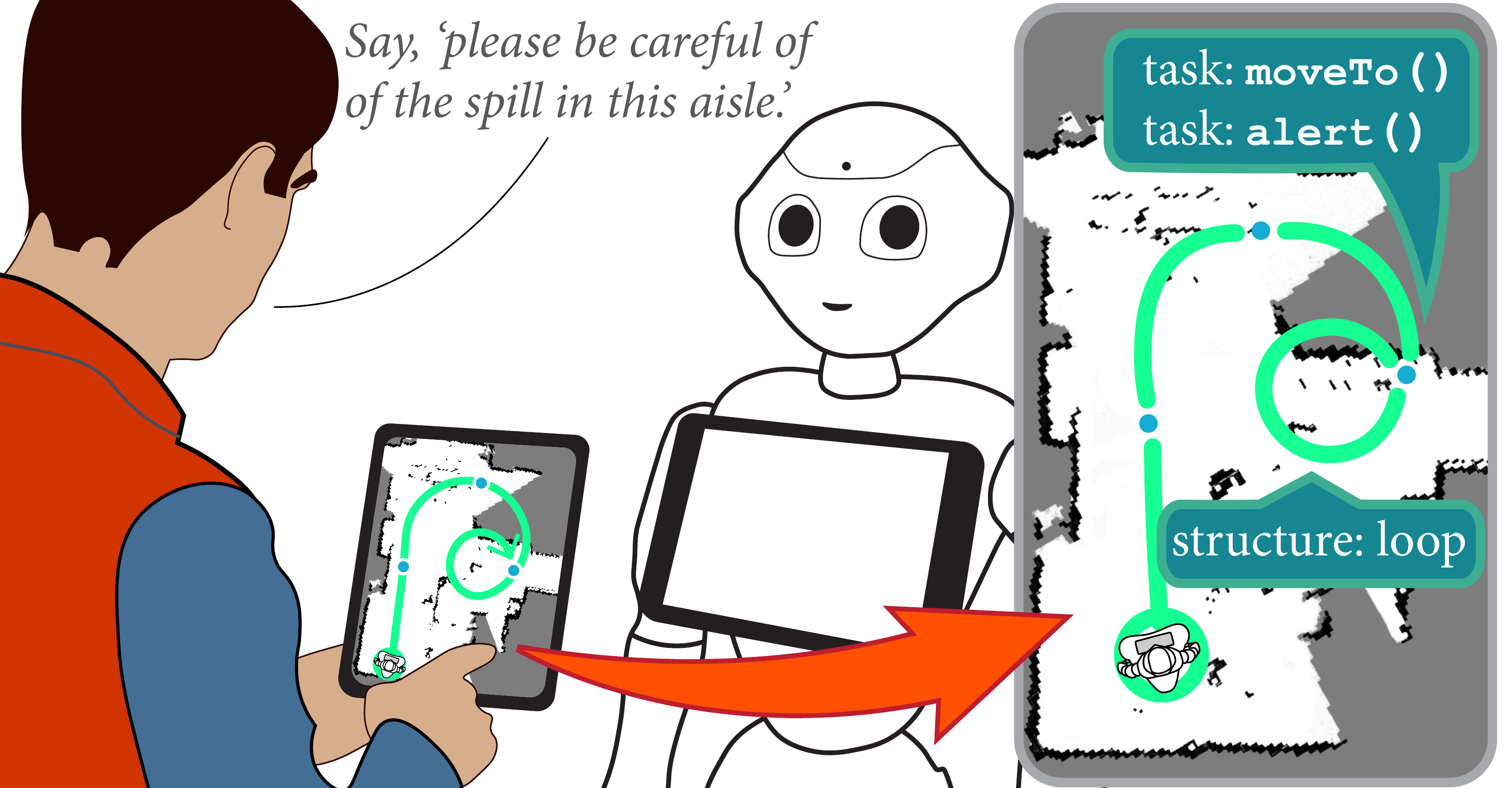}
    \caption{\tool{} lets end-user developers of robots contextualize their speech by sketching out program structure.}
    \label{fig:teaser}
\end{figure}

\section{Introduction}\label{sec:introduction}

\textit{End-user development} (EUD) solutions for robotics must allow end users to easily and efficiently create robot applications to satisfy immediate needs.
Consider an example in which the manager of a grocery store must direct traffic away from a spill in the beverage aisle---a perfect task for a robot to perform and a seemingly simple task to specify.
The manager must direct the robot to the location of the spill while ensuring that the robot avoids the spill; the robot must issue a cautionary statement to anyone approaching the aisle; and the robot must return to its charging station when the spill has been cleaned up (Figure \ref{fig:teaser}).

Although simple in concept, designating the robot's task at a moment's notice may prove challenging. While \textit{learning} techniques promise assistance due to their effectiveness in robotic task training, offline training forgoes critical social and environmental context, while online training takes time. \textit{End users}, by contrast, already possess the contextualized knowledge required to specify a task. Therefore, we posit that programming tools for end users offer a better approach for on-the-fly task specification. Existing tools for robotics, however, are impracticably instrumented for addressing immediate needs (e.g., desktop interfaces),
or demand basic programming knowledge that is tangential to the expertise and skills of domain experts (e.g., automata or block-based programming in \cite{porfirio2018authoring, huang2016design}). Other programming paradigms, in contrast, compensate by restricting expressiveness, such as those that offer simple program representations (e.g., trigger-action programming in \cite{leonardi2019trigger}), or those that limit developers to programming only one aspect of a robot's behaviors (e.g., movement but not task goals in \cite{young2012style}).

To address the on-the-fly programming needs of end-user developers, we created a novel, on-the-fly, EUD solution, \tool{}, designed to reconcile simplicity and expressiveness.
The guiding principle of \tool{} is to capture and automatically refine rapid, incomplete developer input from as minimally instrumented of an interface as possible. Achieving this goal is founded on two design choices. First, a formative design study conducted by the authors and described in this paper suggests that a multimodal interface with partial reliance on speech will enable end users to easily and efficiently express simple tasks for a robot to perform. Second, at present day, touch screen and voice interfaces such as mobile phones, tablets, and smart watches, are ubiquitous. End-user developers can therefore conveniently access touch interaction to contextualize spoken language statements and fill in logic gaps.

Guided by these ideas, \tool{} enables end users to program robots through multimodal speech and sketching input. In a \textit{recording} session, developers utter one or two spoken language 
 statements that correspond to the primary goals, or \textit{core}, of a task. To contextualize the core, developers sketch program logic on a two-dimensional representation of the robot's target environment. When the recording session ends, \tool{}'s program synthesizer leverages automated planning techniques to assemble a task by (1) embedding the robot's goals within the path drawn by the developer and (2) inserting any additional steps required to achieve these goals. If multiple recordings have been provided, the synthesizer combines all the resulting task plans into an executable automaton. 

Our primary contribution is therefore a programming system, \tool{}, and the EUD paradigm that it affords. In this paper, we first describe a formative design study with a speech-only prototype, which ultimately served as a catalyst for the ideation of \tool{}. We then describe \tool{} itself, focusing on the integration of sketching to contextualize speech and specify program logic.

Our contributions are summarized as follows:

\begin{itemize}[nosep]

  \item \textit{System} --- a full-fledged development tool, \tool{}, and a set of application scenarios to demonstrate its use.
  \item \textit{Design} --- a design study that results in design principles for creating on-the-fly EUD tools for robots.
  \item \textit{Technical} --- a program synthesis approach for contextualizing spoken language with program sketches.

\end{itemize}

\section{Related Work}

Our work draws on the literature from end-user development, natural language programming, program synthesis, and planning in artificial intelligence (AI planning). In the following, we briefly discuss relevant key concepts and related work.

\subsection{End-User Development}
End-user development (EUD) aims to democratize programming for novices. \citet{lieberman2006end} characterizes EUD as surpassing application parameterization and customization and allowing users to modify or create programs from scratch. An EUD paradigm of note, trigger-action programming (TAP), has been widely successful in its adoption by end users \cite{ur2016trigger}. However, despite its simplicity, TAP developers are still susceptible to inserting undesirable or unpredictable behaviors into their programs \cite{zhang2019autotap}.
In a different approach to EUD, \textit{sloppy programming} has explored the automatic mapping of coarse text entry to the capabilities of an API \cite{little2011sloppy}. Under the umbrella of the {\em no-code movement}, a recently popularized term for EUD, many commercial products allow users to create complex applications, including \textit{Webflow} for intricate webpages \cite{webflow}, \textit{AirTable} for databases \cite{airtable}, and \textit{Zapier} for automation \cite{zapier}.

Various approaches to EUD have been explored in robotics, but are typically limited in expressive power. These limitations have arisen from restrictive programming paradigms like TAP \cite{leonardi2019trigger,senft2021situated} and input methods like natural language \cite{forbes2015robot}, or from making only a small subset of robot actions available to be programmed (e.g., only motion trajectories as in \cite{young2012style}). More expressive EUD interfaces, however, may increase developer mistakes and compromise the robot's dependability. Prior work in end-user software engineering (EUSE) has sought to preserve dependability by providing end users with standard software engineering practices (e.g., fault localization) \cite{burnett2004end}, and thus may prove useful for robotics. 
Sketching is a familiar concept in robot EUD and control. An especially natural use of sketching involves specifying the navigation path of a robot and the surrounding environment \cite{boniardiSketching} or other navigation-related commands such as a drawn ``X’’ indicating ``go here’’ \cite{shah2010} or a drawn lasso indicating ``vacuum this area’’ \cite{sakamoto2009sketch}. \tool{} draws heavily from \textit{Roboshop}, an interface for annotating a top-down view of a robot's environment with tasks to perform \cite{liu2011roboshop}, and \textit{V.Ra}, a task-authoring interface that integrates navigation paths with both robot actions and program logic \cite{vra}. 
Additionally similar to \tool{} is the work of \citet{shah2012towards} that integrates speech with sketching for specifying navigation commands and the work of \citet{rtcontrol1} and \citet{rtcontrol2} on a real-time robot control interface that also integrates speech with sketching. Among these works, \tool{} derives novelty from its ability to synthesize branching and looping programs from coarse, on-the-fly multimodal input.

\subsection{Natural Language Programming}
Motivated by the widespread use of language in human interactions, researchers have explored several different approaches to allow natural language interactions with robots. Semantic parsing, in which natural language is transformed to a logical representation \citep{woods1973progress,zettlemoyer2005,dong2016language}, has often been used to enable language specification of commands, goals, or simple programs \citep{chen11learning,matuszek2013,thomason2019improving,walker2019neural}. 
Alternative approaches based on syntactic features have also been shown to be effective when matched with appropriate domain knowledge \citep{tellex2011understanding}.
In certain applications, the direct mapping of a controlled subset of English to the target formalism has proven sufficient \citep{kress2008translating}.

Natural language dialogue systems use multi-turn language interactions to better accommodate the communication of complex instructions. In robotics, these systems have enabled end users to specify reusable programs for tasks such as navigation \citep{lauria2002mobile, thomason2019improving}, assembly \citep{stenmark2013natural,forbes2015robot}, and social interaction \citep{gorostiza2011end}. Some have envisioned human-robot dialogue as a way for future domestic robots to acquire necessary environment-specific knowledge, from the actions needed to complete some task to the rules that underlie the world \citep{connell2019verbal}.
In this vein, \citet{mohan2014learning} developed an explanation-based task learning approach, using situated instructions to teach novel hierarchical tasks to a robot, and later work showed how these task representations could generalize across situations \citep{kirk2019learning}.

\subsection{Planning \& Synthesis}

Program synthesis is used to automatically construct fully executable programs from partial developer specifications \citep{gulwani2017program}. In human-robot interaction, program synthesis has been applied in both robot manipulation \cite{gao2019pati, huang2019synthesizing} and social domains \cite{chung2022authoring}. Similar to \tool{}, the programming tool \textit{Figaro} synthesizes robot programs from multimodal speech and touch demonstrations \cite{porfirio2021figaro}. \textit{Figaro}, however, requires developers to recite their speech and touch in the exact order that they must occur in the resulting program.

To synthesize programs, \tool{} uses techniques from AI planning, which is broadly defined by \citet{alterovitz2016robot} as ``computing actions and motions for a robot to achieve a specified objective.'' In accordance with \citet{ghallab2016automated}, we classify our approach as operating at the \textit{descriptive} level, in which plans contain information about what actions for the robot to perform and when to perform them, rather than the \textit{operational} level that describes precisely how the robot should perform these actions. \tool{} draws inspiration from notable successes of planning in human-robot interaction, including the \textit{Human-Aware Task Planner} that plans a robot's actions in accordance with social rules \cite{alili2009task} and work from \citet{petrick2013planning} that plans the actions of a social bartender robot for multi-party human-robot interactions.
\vspace{-6pt}
\section{Speech Prototype: One Mode of Input}\label{sec:proto}

In this section, we describe our prototypical speech interface that ultimately served as a catalyst for the development of \tool{}.\footnote{ Portions of \S\ref{sec:proto} were presented in Chapter 7 of the first author's Ph.D. dissertation \cite{porfirio2022authoring}. \S\ref{sec:proto} focuses on aspects of this work that led to the development of \tool{}. }  In the vein of affording end users as much control with as minimal input as possible, the prototype explores the feasibility of end-user development with a single input modality---\textit{speech}---for two reasons. First, speech is an intuitive form of communication inherent to everyday interaction. Second, sensing speech requires minimal instrumentation (i.e., only  a microphone) beyond the robot itself.

In what follows, we describe the prototypical speech interface, its evaluation, and key lessons that inform \tool{}.
\vspace{-6pt}

\subsection{Prototypical Speech Interface}\label{sec:proto_speech}

The prototype consists of an early version of \tool{}'s verbal input interface consisting of a wakeword recognizer, a speech-to-text engine, and a speech classifier. In the prototype, the verbal interface is accompanied by a simple visual feedback interface.

To use the verbal interface, end users begin designing a task by saying the wakeword ``listen to me.'' Then, users can verbally enter utterances into the interface, each of which is assigned to individual \textit{commands} from an available set---either (1) \textit{action} commands, which specify that the robot must do something, or (2) \textit{event} commands, which specify that the robot should listen for a particular trigger to which the robot can respond, such as someone approaching or speaking. Instantiating event commands is thus the primary means for end users to encode human behavior in a program.  
For the prototype, we developed a small and exploratory set of commands. A sample subset of action (top five) and event (bottom two) commands is listed below:

\begin{center}
\begin{tabular}{ l c l }
 \hline
 \tcmd{moveTo:}{place} & $\longrightarrow$ & move to \textit{place} \\
 \tcmd{put:}{item, place} & $\longrightarrow$ & put the specified \textit{item} in \textit{place} \\
 \tcmd{say:}{speech} & $\longrightarrow$ & say the contents of \textit{speech} \\
 \tcmd{ask:}{speech} & $\longrightarrow$ & ask the contents of \textit{speech} \\
 \tcmd{tell:}{narrative} & $\longrightarrow$ & recite the contents of \textit{narrative}\\
 \tcmd{eventApproach}{} & $\longrightarrow$ & person approaches the robot \\
 \tcmd{eventSpeech:}{speech} & $\longrightarrow$ & person says \textit{speech} to the robot\\
 \hline
\end{tabular}
\end{center}

For the remainder of the paper, we refer to a \textit{command} as a fully instantiated action or event in which all parameters in the command are resolved. A command \textit{type} refers to an uninstantiated command. For example, the type of \tcmd{say:}{`hello'} is \textbf{\texttt{say}}, while the parameter of the instantiated command is \textit{`hello.'}

To infer a command from an utterance, the prototype uses a non-learned, keyword-based approach that scores commands based on how well verbs and nouns in the utterance match a command's type and parameters, respectively. Scores are derived by querying keywords---verbs, nouns, command types, and parameters---within WordNet \cite{wordnet1995,wordnet1998} and extracting the real-value distances between synonyms of these keywords. For example, within the utterance, ``Put the groceries in the kitchen,'' the action command \tcmd{put:}{groceries, kitchen} scores highly because the words ``put,'' ``groceries,'' and ``kitchen'' match the command type and parameters. 

Event commands score higher than action commands if the utterance contains keywords like ``if'' or ``when.'' For example, in the utterance ``When someone says `hello,'\thinspace'' the event command \tcmd{eventSpeech:}{`Hello'} (someone greets the robot) scores higher than its corresponding action command \tcmd{say:}{`Hello'} (the robot says ``hello'') because the utterance begins with the word ``when.'' 

For speech commands, we require that the user provide the exact speech that the robot should utter or the exact speech that the robot can recognize. For example, if the end user wishes to specify that the robot emits a greeting, the user can say something like ``The robot should now say `\textit{Hello, it's nice to see you!}'\thinspace''  in order to produce the corresponding action command \tcmd{say:}{`Hello, it's nice to see you!'}

As end-user developers produce a sequence of utterances, the prototype produces a program that consists of the corresponding sequence of commands. The sequence of commands is displayed on the visual feedback interface for the user to check. For editing in-progress command sequences, the prototype contains three simple directives: ``undo'' for undoing commands, ``redo'' for redoing commands, and ``reset'' for deleting all commands and starting over.

\subsection{Formative Evaluation}\label{sec:protone_eval}

To evaluate our design decisions within the prototype, we conducted a remote user study over separate video calls with five participants (three males, two females) aged 18 to 43 years ($M=24$, $SD=10.7$). Participants had little to no experience with robots and mixed levels of programming experience. The study was approved by an institutional review board (IRB).

In the study, participants were trained to use the prototype and presented with three tasks within a simulated home environment (e.g., welcoming someone home). 
For each of the three selected tasks, participants were allotted three minutes to program the robot and test the robot in a low-fidelity  simulator within which participants could  execute their programs over the video call. In the test environment, an icon of a robot moved around the home and interacted with participants via microphone and speaker.

At the end of the study, we asked participants to respond to the System Usability Scale \textit{SUS} (10 items on a five-point rating scale) \cite{brooke1996sus} and the \textit{USE} questionnaire \citep{lund2001measuring}, which measures usefulness, ease of use, ease of learning, and satisfaction (30 items on a seven-point rating scale). The prototype's average \textit{SUS} score was 77 $(SD=17.9)$. Within \textit{USE}, on a scale of one to seven,  participants rated the prototype's usefulness 4.7 $(SD=1.85)$, ease of use 4.45 $(SD=1.82)$, ease of learning 4.85 $(SD=2.22)$, and satisfaction 4.31 $(SD=2.15)$. 

Additionally, we conducted brief (5-10 minute) semi-structured interviews with participants to obtain a richer understanding of their experience. To analyze the interviews, we performed open coding and extracted key themes from the interview data, which we summarize below. Participants are referred to as P1-5.

\begin{itemize}
    \item \textit{Theme 1, on-the-fly task specification}: Despite issues with speech recognition and classification (see below), the interface was viewed as intuitive and easy by some participants (P2, P4, P5). P1 even described its potential to be "on the fly." In line with the guidelines proposed by \cite{amershi2019guidelines}, however, P1 highlighted the need to modify programs after their creation.
    \item \textit{Theme 2, shortcomings of spoken language}: Participants expressed difficulty with spoken language, stating that it was ``clunky and inefficient'' (P2) and required them to adjust their speech style (P5). P3 additionally highlighted the ambiguities inherent in speech, such as being unclear who is referring to whom when involving other people in a task.
    \item \textit{Theme 3, preferences on specification paradigm}: While linear task specification was viewed favorably (P4), P3 expressed that the interface was ``too simple,'' and P1 and P4 highlighted potential insufficiencies in the interface in handling complex programs. Other participants preferred alternative input methods such as through ``typing'' (P2, P4), a ``Scratch''-like interface \citep{resnick2009scratch} (P1), or customizable ``block'' commands (P2).
\end{itemize}

In addition to our questionnaire and interview data, we observed various \textit{usage patterns} that help characterize how participants used the prototype. Participants required an average of $67.4$ seconds ($SD=29.6$ seconds) to specify each task. One participant did not finish one task, so we excluded this task from the average specification time. We observed participants experience difficulty with the speech interface due to a combination of incorrect speech transcription, speech not being heard altogether, the wakeword not being recognized, and speech being misclassified even if heard correctly. Possibly due to these difficulties, participants used the ``undo'' or ``reset'' directives an average of $1.47$ ($SD=1.30$) times per task.

\subsection{Implications of Prototype}
We now discuss the implications of the prototype that emerged from our study as they pertain to the current version of \tool{}.

\textbf{Speech. }  
While the prototype was viewed favorably as an on-the-fly tool (\textit{Theme 1}), participants expressed difficulties with the speech interface (\textit{Theme 2}), as further evidenced by our objective observations of these difficulties and the number of times that participants used the ``undo'' and ``reset'' directives (\textit{usage patterns}). We thereby determined that \tool{} \textit{should reduce its reliance on speech and provide support for underspecifying verbal commands.}

\textbf{Additional modes of interaction. } In response to participant feedback (\textit{Theme 3}) and to compensate for the reduction in speech, we determined that \textit{\tool{} should afford users with a second input channel that (1) helps infer task details without requiring the user to specify these details verbally and (2) allows end users to more effectively understand and manage potential program complexity.} This input channel should avoid the need for additional instrumentation (e.g., requiring a keyboard and mouse).
\section{\tool{}: Two Modes of Input}\label{sec:technical_approach}

\begin{figure}[!t]
  \centering
  \includegraphics[width=\columnwidth]{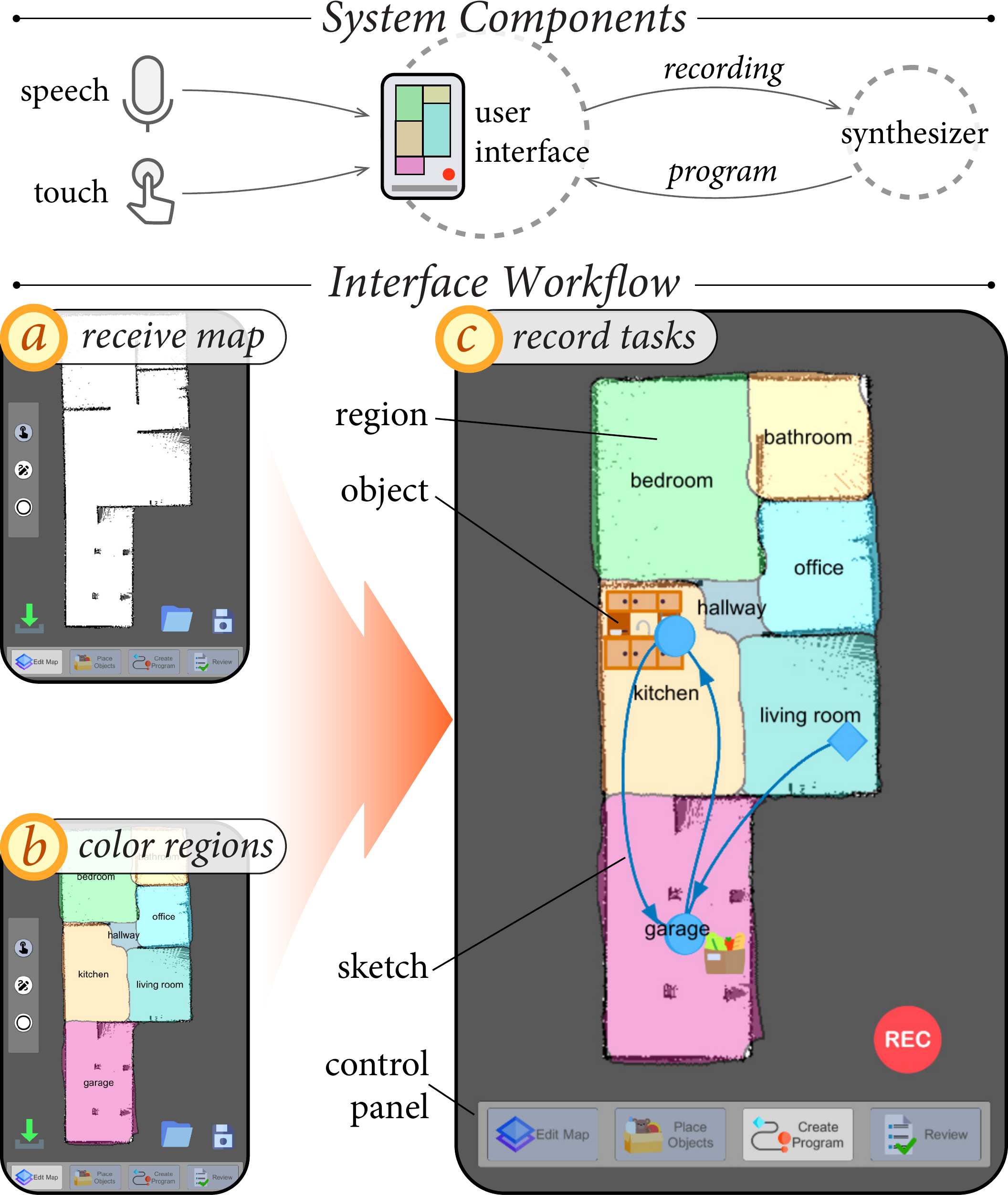}\vspace{6pt}
  \caption{(Top) \tool{} consists of an interface that passes recordings to a synthesizer, which returns a program. (Bottom) The interface enables users to (a) receive a map from a robot, (b) color important map regions, and (c) record tasks.}
  \label{fig:interface}
\end{figure}

The design study with the prototype illuminates various challenges that end-user developers face with speech as their sole input modality. Due to the implications of the study, in addition to prior work that highlights various benefits of multimodal interfaces (e.g., inclusiveness and accessibility \cite{worsley2018multimodal}), we supplemented speech with an additional modality, sketching, to create \tool{}. We chose sketching in order to deemphasize speech by enabling end users to tactilely contextualize a small set of \textit{core}, possibly underspecified commands. Sketching further allows end-user developers to craft program logic (e.g., loops) that are difficult to express verbally and maintains our goals of requiring minimal instrumentation for developers to complete a development task.

The \tool{} system is implemented within two components communicating over ROS Noetic\footnote{\url{http://wiki.ros.org/noetic}}---a handheld touch or stylus-based \textit{interface} implemented in Unity version 2020.3.21f1\footnote{\url{https://unity.com/}} and a \textit{synthesizer} implemented in Python 3 (Figure \ref{fig:interface}, top). Given a two-dimensional map of the robot’s environment (Figure \ref{fig:interface} bottom, a) with labelled regions (Figure \ref{fig:interface} bottom, b), users verbalize a set of core commands and sketch the intended path of the robot on the map (Figure \ref{fig:interface} bottom, c). Subsequently, the interface sends the  recording consisting of the user's speech and sketch to the synthesizer, which returns a program to the interface.\footnote{Implementation of \tool{} took place at the University of Wisconsin–Madison. Code and test cases for \tool{} are available at \url{https://github.com/Wisc-HCI/Tabula}. Additional auxiliary material is available at \url{https://osf.io/jktph/}.}

In what follows, we describe (1) how \tool{} is configured for use, (2) how users then create \textit{recordings} from speech and sketches, and finally (3) how programs are synthesized from recordings.

\subsection{Getting Ready to Use \tool{}}\label{subsec:motivating_example}

Consider the following motivating example: a user wishes to program a robot to meet them every time they return from grocery shopping to help with unloading. In order to use \tool{}, technical requirements must be satisfied, i.e., provide underlying assumptions of the robot's capabilities and populate a map to use with \tool{}.

\textbf{Robot Assumptions. } The developer must have access to a robot that is capable of creating a two-dimensional map of its environment, within which it should be able to accurately localize itself and recognize objects. 
In addition, the robot must be equipped with state-of-the-art path, motion, and task planners---it should be able to navigate to different areas in the environment, interact with objects that it recognizes, and handle edge cases in its task within reason (e.g., if the robot has a goal to grab groceries but the groceries are inside of the user's car, the robot will know to open the car door and search for the groceries before grabbing them).

\textbf{Knowledge Handling. } Prior to use, \tool{} must possess contextual knowledge. Knowledge handling within \tool{} draws heavily from prior work in AI planning, particularly \citet{petrick2013planning}, in that \tool{} contains a fixed \textit{domain} that describes the universe of known possible entities that the robot is assumed to be able to recognize and interact with (e.g., types of objects and humans), the semantics of each entity (e.g., ``cabinet'' is a ``container''), a set of available commands that consist of actions for the robot to perform or events that it should wait for, and preconditions that must be met to perform or post-conditions that hold true as a result of some commands. Also in accordance with common practice, \tool{} stores current world state within a dynamic, modifiable \textit{world} database. 

\textbf{Map Setup. } Prior to specifying a task within the robot's environment, end users may use \tool{} to request the robot's most up-to-date two-dimensional map  (Figure \ref{fig:interface} bottom, a).
Then, \tool{} is used to color \textit{regions} of interest, or areas on the map that the robot is expected to visit (Figure \ref{fig:interface} bottom, b). Finally, the user can use the interface to add objects to the map that may also be of interest to the robot. For instance, the user may place a ``groceries'' icon in the garage region, thus adding it to the \textit{world} database and indicating to the robot that it can find groceries in the garage. The latter step of placing objects in regions is not a strict requirement.

\subsection{Recording a Task}

When an end user is ready to program their robot, they create a \textit{recording}, shown in Figure \ref{fig:synthesis}a. A recording consists of one utterance \utt{} and one sketch \skt{}. The utterance is intended to describe the core of the task for the robot to perform, while the sketch is intended to ground the utterance  within the robot's surrounding environment.

Using our motivating example for illustration, when the end-user developer is ready to embark on their shopping trip, they pull out their phone, activate the \tool{} app, and press the ``Record'' button. While recording, the end user's first action is to verbalize the task core: \textit{``when I arrive, bring in the groceries.''} \tool{} uses the Stanford CoreNLP library \cite{manning2014stanford} to detect \textit{``when I arrive''} as a subordinate clause and splits the user's speech accordingly into two separate parts. Then, \tool{} parses each clause into individual commands, shown in Figure \ref{fig:synthesis}a-b, using a similar approach to \S\ref{sec:proto_speech} with a few notable differences. First, \tool{} foregoes a scoring-based approach in favor of pure keyword matching to map nouns in \utt{} to command parameters and VerbNet \cite{kipper2000class} (rather than WordNet) to map verbs in \utt{} to synonyms of command types. 
\tool{} also supports partially specified commands, such as commands that contain unfilled parameters. Given these modifications, \tool{} parses \textit{``when I arrive''} to a candidate event command \tcmd{eventApproach}{} and \textit{``bring in the groceries''} to a candidate action command \tcmd{put:}{groceries, \_\_\_\_}, in which the blank line represents an unspecified argument. 

Occurring either before, during, or after verbalizing the task core, the end user sketches the sequence of regions that the robot should visit. Beginning in the living room region, the developer slides their finger to the garage region, then to the kitchen region, and then back to the garage. \tool{} parses the sketch \skt{} into the sequence of regions $\mathit{garage} \rightarrow \mathit{kitchen} \rightarrow \mathit{garage}$, omitting the first location (\textit{living room}) so as not to restrict the robot to begin its task in any one region on the map. Figure \ref{fig:synthesis}a-b depicts the step of parsing \skt{} to a region sequence. 

\subsection{Program Synthesis and Output}\label{sec:synthesis}

\begin{figure}
  \centering
  \includegraphics[width=\columnwidth]{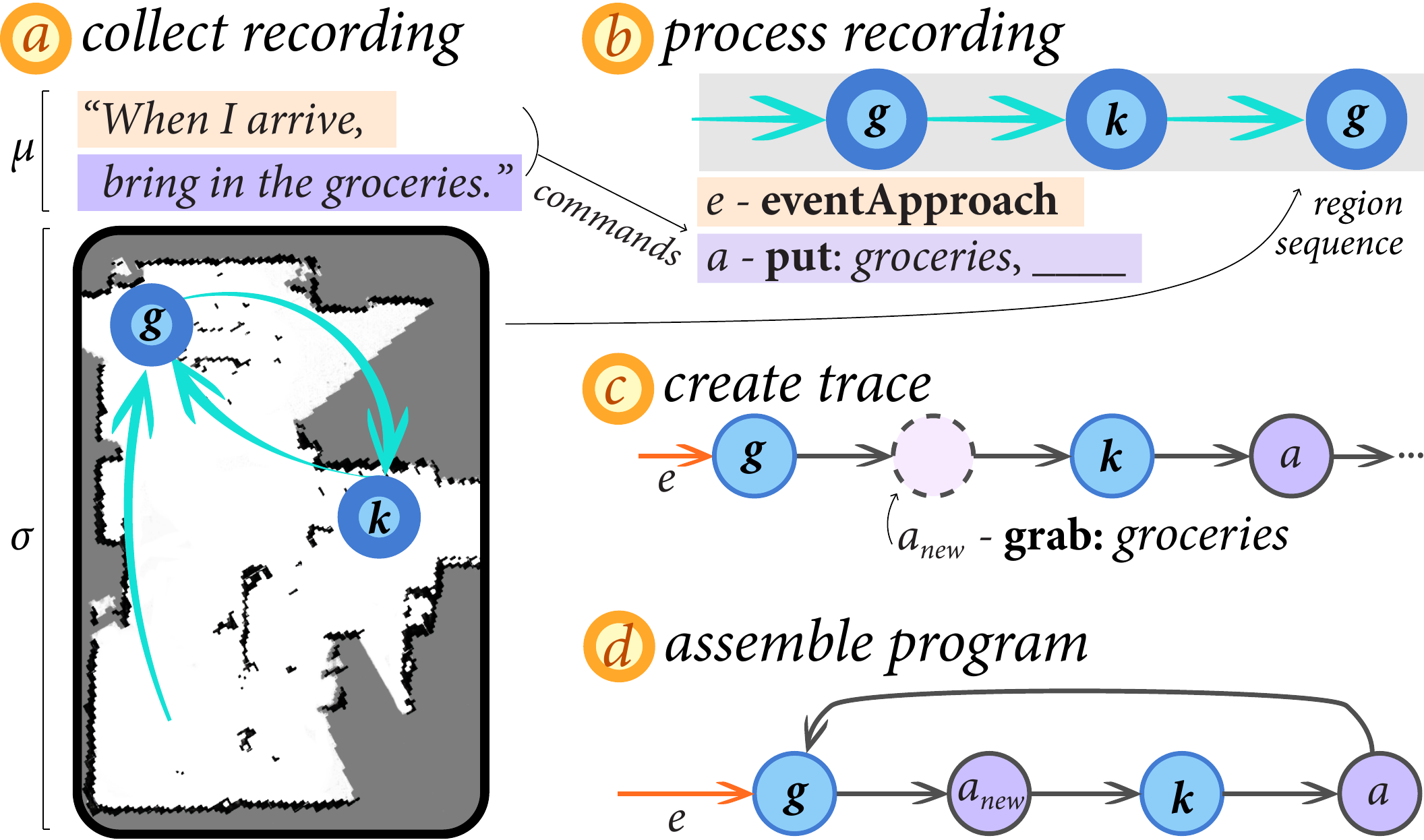}\vspace{6pt}
  \caption{The technical approach for program synthesis from a lone recording, where \textit{e} refers to \textit{event} and \textit{a} refers to \textit{action}.}\vspace{-6pt}
  \label{fig:synthesis}
\end{figure}

Given one or multiple recordings provided by the end user, the goal of the synthesizer is to (1) create \textit{traces} from each recording and (2) assemble a finite automaton, or \textit{program}, that accepts each trace.

\textbf{Creating a Trace from a Recording. } Given a recording \rec{} containing a parsed utterance \utt{} and parsed sketch \skt{}, the synthesizer must create a trace \trc{} that satisfies the constraints set by parsing \utt{} and \skt{}. A trace is a sequence of robot actions $\act{}_0\xrightarrow{\con{}_0}\act{}_1\xrightarrow{\con{}_1}\dots\xrightarrow{\con{}_{n-1}}\act{}_n$ where $\act{}_i$ is the $i$th robot action and $\con{}_i$ is the $i$th event. Figure \ref{fig:synthesis}c illustrates the task of formulating a trace \trc{} from individual components \utt{} and \skt{}. 

To illustrate, recall our example with the utterance \textit{``when I arrive, bring in the groceries''} and the sketch from the living room to the garage, to the kitchen, and then back to the garage. For clarity in describing how \tool{} creates a trace for this recording, let us begin by considering a simpler example in which the garage is visited only once (we will return to our full motivating example in \S\ref{sec:synthesis}, \textit{Loops}). The utterance is still parsed to the commands \tcmd{eventApproach}{} and \tcmd{put:}{groceries, \_\_\_\_}, but the sketch is parsed to the shortened sequence of regions $\mathit{garage} \rightarrow \mathit{kitchen}$. With our shortened sketch, the task of the synthesizer is to create trace \textit{t} as follows, where unlabeled transitions refer to the empty event in which the robot needs no prompting to perform one action after another: 

\vspace{-9pt}
\begin{equation*}
\begin{aligned}
&\tcmd{idle}{}\xrightarrow{\tcmd{eventApproach}{}}\tcmd{moveTo:}{garage}\xrightarrow{}\tcmd{grab:}{groceries}\xrightarrow{} \\
&\tcmd{moveTo:}{kitchen cabinets}\xrightarrow{}\tcmd{put:}{groceries, kitchen cabinets}
\end{aligned}
\end{equation*}

In its search for trace \trc{}, the synthesizer must make multiple decisions autonomously: (1) within which regions the core commands from \utt{} should be inserted, (2) how to resolve unfilled arguments from these core commands, (3) whether and where additional robot actions need to be inserted such that the preconditions of each command in the trace are satisfied, and (4) whether and how the \textit{world} database needs to be modified such that the robot can complete the trace successfully. In order to make these decisions, the synthesizer employs A* search to plan for the most optimal trace in terms of discrete actions and locations. The planning space includes the following penalties:

\begin{enumerate}
  \item Traces incur penalties equal to their length. Longer traces are thus more costly than shorter traces.
  \item Each region or entity that the robot visits incurs an additional penalty if the robot does no action at that location.
  \item Any entity that exists in the trace but has not yet been inserted in the \textit{world} incurs an additional penalty.
\end{enumerate}

The planning space includes the following additional constraints: the synthesizer will only accept traces that (1) include \tcmd{moveTo}{} commands for each region present in the original sketch, and (2) include the core commands specified by the end user's utterances. If an object exists in an accepted trace that does not yet exist within \tool{}'s most up-to-date snapshot of the robot's environment (the \textit{world} database), the object will be added to the \textit{world} database.

To illustrate the planning space within our shortened example, the synthesizer makes the following decisions. The \tcmd{eventApproach}{} core command is inserted before the robot moves to the garage and the core \tcmd{put:}{groceries, \_\_\_\_} command is inserted when the robot is in the kitchen. In deciding how to resolve the \tcmd{put:}{groceries, \_\_\_\_} command with the unfilled argument for where the robot should place the groceries, the synthesizer searches for an entity in the domain labelled as ``container'' and existing in the kitchen region, and completes the command with the argument \textit{kitchen cabinets}. In determining whether and where additional robot actions are needed in \trc{}, the synthesizer knows from the planning domain that a precondition of \textbf{\texttt{put}} is that the robot must first be holding an entity before it is able to put it somewhere. Therefore, the synthesizer decides to insert a \tcmd{grab:}{groceries} command for when the robot is in the garage. Lastly, if the \textit{world} database does not already indicate that \textit{groceries} can be found in the garage, the synthesizer will modify \textit{world} accordingly and incur a penalty.

\textbf{Assembling a Program. }
Given a single trace, there may be nothing left for the synthesizer to do -- the trace itself becomes a step-by-step program for the robot to execute. If the developer inserts loops into a recording or provides multiple recordings, then the synthesizer will have additional work. 

\textit{Loops. }Within a single recording, end users may introduce loops. To do this, end users need only visit a region multiple times in the course of a sketch. To illustrate, let us return to our original motivating example in which the recording still consists of \utt{} \textit{``when I arrive, bring in the groceries''} and \skt{} once again consists of the robot moving from the living room to the garage, from the garage to the kitchen, and then back from the kitchen to the garage. As before, \skt{} is parsed to the sequence of regions $\mathit{garage} \rightarrow \mathit{kitchen} \rightarrow \mathit{garage}$.

The synthesizer detects a loop within the sketch (\textit{garage} is repeated) and then extends the loop such that there are two iterations total and each loop iteration is identical. Taking into account $\mathit{garage} \rightarrow \mathit{kitchen}$ as a single loop iteration, the sketch will be extended so that this iteration completes twice, producing the following modified sketch \sktloop{}: $\mathit{garage} \rightarrow \mathit{kitchen} \rightarrow \mathit{garage} \rightarrow \mathit{kitchen}$.

For producing a trace from \utt{} and \sktloop{}, an additional synthesis constraint is necessary---for any location (i.e., a region or entity) visited multiple times in a trace, the sequence of actions and events occuring at that location must always be the same. The resulting trace is therefore as follows:

\vspace{-9pt}
\begin{equation*}
\begin{aligned}
&\tcmd{idle}{}\xrightarrow{\tcmd{eventApproach}{}}\tcmd{moveTo:}{garage}\xrightarrow{}\tcmd{grab:}{groceries}\xrightarrow{} \\
&\tcmd{moveTo:}{kitchen cabinets}\xrightarrow{}\tcmd{put:}{groceries, kitchen cabinets}\xrightarrow{} \\
&\tcmd{moveTo:}{garage}\xrightarrow{}\tcmd{grab:}{groceries}\xrightarrow{} \\
&\tcmd{moveTo:}{kitchen cabinets} \xrightarrow{}\tcmd{put:}{groceries, kitchen cabinets}
\end{aligned}
\end{equation*}

To assemble the final program, the synthesizer combines repeated sequences of actions and conditionals to form a loop, shown in Figure \ref{fig:synthesis}d. 

\textit{Multiple Recordings. }
After an initial recording has been provided, additional recordings can be \textit{attached} to any existing recording. Attached recordings cannot start from any arbitrary location in the world; rather, they must branch from a location within an existing sketch. The synthesizer assembles traces from each recording no differently than if only one recording was provided.

Assembling an executable program from an initial trace and one or more \textit{attached} traces is straightforward. If the end user begins an attached recording at a location $l$ with a core \textit{event} command (i.e., ``when I say `stop helping me with the groceries'\thinspace''), the resulting program will contain a branch at $l$ in which the trace resulting from the attached recording will execute immediately when the event occurs. It is possible for nondeterminism to arise from the attachment of traces to each other, such as if the end user begins an attached recording without providing a core event command.
\section{\tool{} Capabilities and Limitations}\label{sec:app_scenarios}
We demonstrate \tool{}'s capabilities by describing a set of application scenarios. Next, we utilize a suite of 33 total synthesizer test cases that cover, but are not limited to, different variations of these scenarios in order to provide an analysis of \tool{}'s reliability.

\subsection{Application Scenarios}

\begin{figure}
  \centering
  \includegraphics[width=\columnwidth]{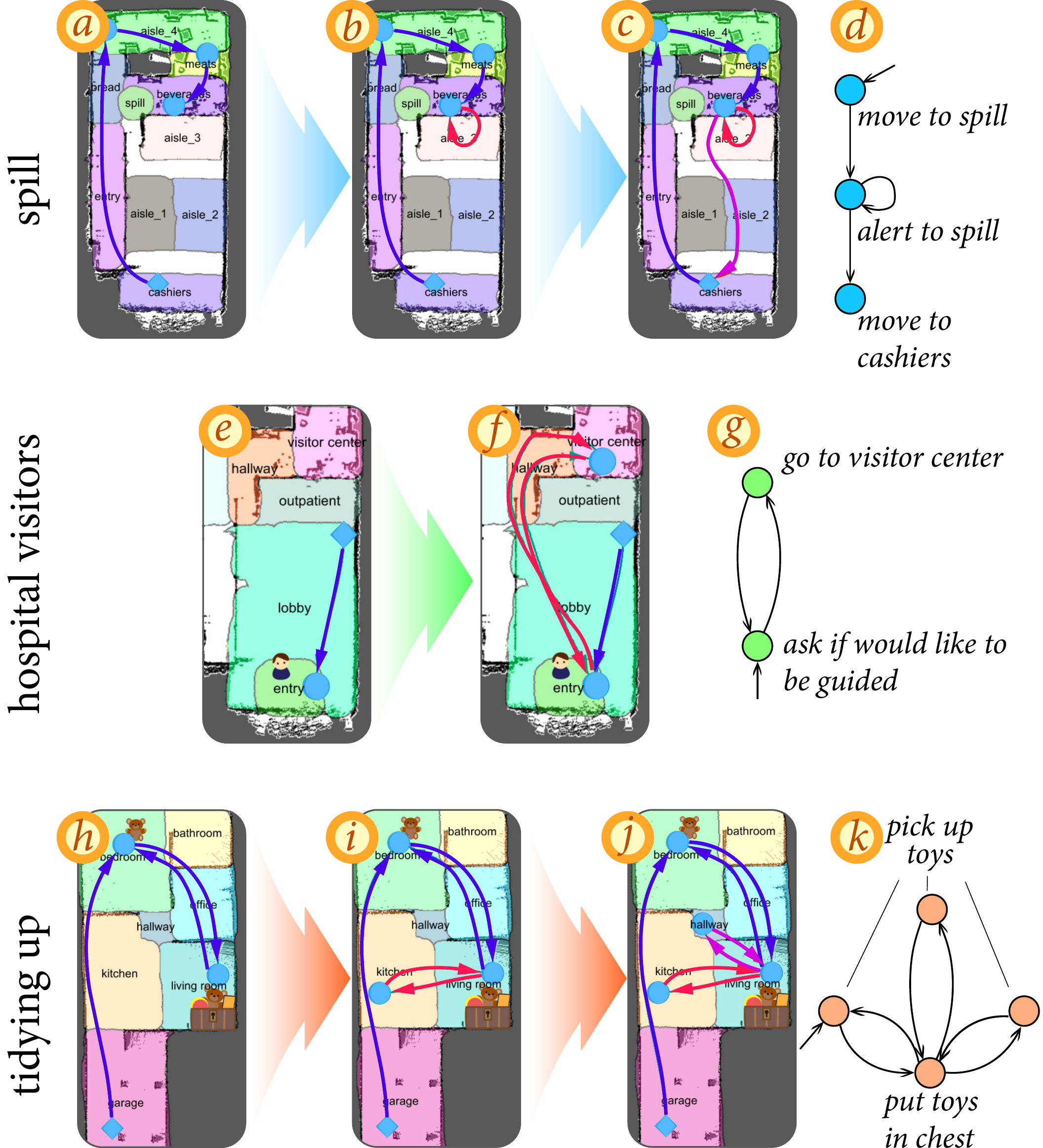}\vspace{6pt}
  \caption{Our application scenarios include alerting people to a spill (top), guiding people to the visitor center in a hospital (middle), and tidying up after a playdate (bottom). Drawn sketches are graphically enhanced for clarity.}\vspace{-6pt}
  \label{fig:application_scenarios}
\end{figure}

In addition to the \textit{Grocery} scenario introduced in \S\ref{sec:technical_approach}, we demonstrate \tool{}'s capabilities with three additional application scenarios. 

\textbf{Alerting People to a Spill. }
This scenario is identical to the one introduced in \S\ref{sec:introduction} where the manager of a grocery store needs to direct traffic away from a spill in the beverage aisle. Recall the task requirements that the robot must move to the location of the spill while avoiding the spill, issue a cautionary statement to anyone approaching the aisle, and return to its starting point after the spill is cleaned up. This application scenario demonstrates on-the-fly task contextualization and using \tool{}'s branching and looping functionality to create trigger-action programs.

Figure \ref{fig:application_scenarios} (top) depicts the steps taken by the manager to program the robot. First, the manager contextualizes the task by drawing a new region to indicate where the spill occurred. Next, the manager creates a recording to direct the robot from its starting point to the beverages, purposefully circumventing the spill so that the robot avoids driving through it (Figure \ref{fig:application_scenarios}a). To ensure that anyone who enters the robot's vicinity is alerted to the spill, the manager then sketches a self-loop in the beverages aisle and utters ``When someone approaches the aisle, say, `Please avoid the spill in this area. It will be cleaned shortly''' (Figure \ref{fig:application_scenarios}b). The effect of this recording is to create a trigger-action program that remains in effect while the robot is in the beverages aisle---whenever someone gets near the robot, the robot will alert them to the spill. Finally, the manager directs the robot back to its starting point by sketching a trajectory from the beverages aisle back to the cashiers and uttering ``When I say `go home''' (Figure \ref{fig:application_scenarios}c).  Figure \ref{fig:application_scenarios}d presents a decontextualized, high-level illustration of the resulting program.

\textbf{Guiding Visitors in a Hospital. }
Consider an employee at a busy hospital wing who wants to streamline the check-in process for visitors. The robot should offer to escort visitors from the hospital entrance to the visitor center. This application scenario is intended to highlight how human behavior can be encoded into a program.

Figure \ref{fig:application_scenarios} (middle) depicts the steps taken by the employee. First, the employee inserts a \textit{person} entity in the entrance to the hospital, indicating to the robot that it will encounter people in this area. The employee then sketches a path from the entrance to the visitor center and utters, ``Tell people the directions to the visitor center. Say, `would you like me to escort you there?'\thinspace'' (Figure \ref{fig:application_scenarios}e). Next, the employee utters, ``If they say `yes,'\thinspace'' and in response to the robot hearing ``yes,'' sketches a path from the entrance to the visitor center and back to the entrance (Figure \ref{fig:application_scenarios}f). The resulting program, depicted in a decontextualized and high-level form in Figure \ref{fig:application_scenarios}g, will thereby loop forever in which the robot approaches people in the hospital entrance, asks them if they are interested in being escorted, and if so, escorts them to the desk.

\textbf{Tidying Up. }
Consider a parent with toys scattered around their home after a playdate. The parent wants to program the robot to remove toys from three specific rooms in their home and place the toys in a chest in the living room. This application scenario is intended to demonstrate looping tasks and the synthesizer's ability to place new objects in a scene.

Figure \ref{fig:application_scenarios} (bottom) depicts the steps taken by the parent. The parent begins by uttering, ``Put the toys in the chest,'' and sketching a loop from the robot's starting point to the bedroom, the living room, and then back to the bedroom. Based on this input, \tool{} inserts a toy object into the bedroom and a toy chest object into the living room (Figure \ref{fig:application_scenarios}h).
The user then provides the same utterance and sketches a path from the chest, to the kitchen, and back to the chest (Figure \ref{fig:application_scenarios}i). Finally, the parent provides the same utterance a last time while directing the robot to the hallway (Figure \ref{fig:application_scenarios}j). In the resulting program, depicted in a decontextualized, high-level form in Figure \ref{fig:application_scenarios}k, the robot loops on picking up toys from the bedroom, hallway, and kitchen until no toys remain.

\subsection{Synthesizer Reliability}
Our suite of 33 synthesizer test cases (referred to as T1-33) allows for a high-level analysis of \tool{}'s reliability. Each test case provides the following input to the synthesizer: (1) the end-user's speech, (2) the end user's sketch, and (3) a custom \textit{world} database tailored to the test case. Given this input, the synthesizer produces a program as output, taking an average of 3.04 ($SD=0.70$) seconds per test case on an Intel Core i7-1065G7 CPU (1.30 GHz). Based on our experiences from constructing our test suite, we have observed three categories of reasons for which the synthesizer might fail to produce the intended output, which we detail below.

\textbf{Insufficient Information from Speech. } While \tool{} is robust to omissions from end-user speech, failure to synthesize the intended program may occur if supporting information is also missing from the \textit{world} database. Consider the \textit{Groceries}  scenario presented in \S\ref{subsec:motivating_example}, encapsulated in test case T8. Had the end user been vague in their speech (e.g., ``bring them in,'' rather than ``bring in the groceries''), T8 will still produce the correct output if the groceries entity is present in the \textit{world} database. However, if the groceries entity is missing both from user speech and the \textit{world} database, the synthesizer will nondeterministically choose an item from the domain to insert into the world for the robot to grab, which may not be groceries.

\textbf{Insufficient Information from Sketching. } Although much information about the robot's core task is provided through speech, sketching provides contextualization of the task within the robot's environment and information about program structure (i.e., branching and looping). Consider the \textit{Hospital} scenario presented in \S\ref{sec:app_scenarios} (T16). The scenario contains a loop in which the robot proceeds back to the entrance after escorting a visitor to check in. If the developer does not explicitly sketch the path back to the entrance, this loop will not be inserted in the program.  

\textbf{Insufficient Domain Knowledge. } This category of failure pertains to the synthesizer not possessing enough prior domain knowledge for contextualization. For example, consider the \textit{Tidying} application scenario presented in \S\ref{sec:app_scenarios} (T26). Instead of cycling between picking up a toy and dropping it in the chest, perhaps the end user wants the robot to first collect all toys, and then drop them into the chest at once. The end user may correctly sketch a path through the kitchen, bedroom, and living room and say ``pick up the toys.'' In this case, however, the synthesizer does not possess enough prior knowledge about the different ways in which tidying can be performed and without this information directs the robot to pick up toys from a single room rather than \textit{each} room. 

\vspace{-8pt}
\section{Discussion}

\subsection{On-the-Fly End-User Robot Development}
There is a need for tools that enable end-user developers to rapidly and conveniently script robot programs for situations that arise spontaneously. 
Although robots are well-suited to handle these situations, development solutions that afford meticulously crafting highly contextualized applications may be difficult and slow to use. In contrast, hands-off techniques stemming from machine learning are highly effective at generating and refining robot applications, but the offline application of these techniques results in decontextualized task specifications, while the online application of these techniques in the intended interaction context requires arduous data collection. With \tool{}, we posit that the best way to rapidly obtain contextualized information about a task at hand is through simple forms of input from end users themselves, who represent domain experts within the robot's target context.

We believe that \tool{} represents a significant step in this direction.
First, \tool{} is quick to use. As demonstrated by our application scenarios and test cases, a full application can be developed using, at minimum, a single speech utterance paired with a single sketch. Furthering its versatility, \tool{} requires very little instrumentation other than a robot and a personal mobile device. 
Furthermore, \tool{} enables task contextualization, owing to the ability to customize the robot's environment and ground spoken language commands within this environment. Users can therefore apply \tool{} to create a robot program in any situation in which the robot is able to localize within its environment. 
Lastly, \tool{} handles complexity without requiring users to pore over task details. With a few simple spoken language commands and the high-level program logic derived from the user's sketch, \tool{} synthesizes a finite state automaton. 

\vspace{-6pt}
\subsection{Limitations \& Future Work}

A key limitation of \tool{} is its lack of evaluation with potential robot end users. As such, we cannot conclude whether \tool{} is more effective than existing state-of-the art solutions for scripting contextualized robot applications on the fly, and we cannot offer conclusive design implications for how \tool{} may be improved. Plans for additional data collection are therefore underway. 

Second, although \tool{} is intended for non-programmers and technical non-experts, we still believe that end users must be trained on how to use \tool{} to its full potential. In particular, we expect that forming a loop within a single recording or creating a branching program from multiple recordings will require practice. Furthermore, minimal training may be required for end users to learn how to optimally specify goals via natural language. Conducting a qualitative, exploratory evaluation of \tool{} will enable us to understand precisely where training is required.

Third, \tool{} lacks in offering feedback to end users and the ability to refine and correct programs, both of which are critical to successful human-AI systems \cite{amershi2019guidelines}. Future work must first provide end users with information about potential faults or unexpected program behavior, such as branches with underspecified triggering events, and then provide end users with the means to correct these issues. Correcting issues will necessitate expanding \tool{}'s refinement capabilities, such as by allowing end users to target and fine-tune specific aspects of a recording.

Fourth, as described in \S\ref{subsec:motivating_example}, \tool{} requires domain knowledge, including a map of the environment, prior to task specification. While \tool{} already somewhat challenges this requirement---environment mapping may be achieved during, rather than prior, to task specification if the end user sketches paths to unmapped areas---interacting with entities not already in the domain is not yet supported. Future work on \tool{} should integrate auto-classification of novel entities within \tool{} or modify \tool{}'s user interface to prompt the end user to classify these entities for the robot.
\vspace{-4pt}

\section{Conclusion}
We present \tool{}, a system for on-the-fly end-user development of robot programs. \tool{} is motivated by the need for simple programming interfaces that maintain the expressiveness required for robot development. We thereby approached the design of \tool{} from the ground up, beginning with an initial speech-only prototype. Based on the results of a design study, we created \tool{} to supplement speech with an additional mode of input, \textit{sketching}. In a series of application scenarios, we demonstrate how \tool{} can create meaningful robot programs through speech and sketching.

\begin{acks}
This work was supported by the National Science Foundation (NSF) award 1925043, 
NSF Graduate Research Fellowship Program award DGE-1747503, and a Cisco Wisconsin Distinguished Graduate Fellowship. This work was carried out while DP was affiliated with the University of Wisconsin–Madison and completed while an NRC Postdoctoral Research Associate at the Naval Research Laboratory.
\end{acks}

\bibliographystyle{ACM-Reference-Format}
\balance
\bibliography{paper}


\begin{thebibliography}{56}


\ifx \showCODEN    \undefined \def \showCODEN     #1{\unskip}     \fi
\ifx \showDOI      \undefined \def \showDOI       #1{#1}\fi
\ifx \showISBNx    \undefined \def \showISBNx     #1{\unskip}     \fi
\ifx \showISBNxiii \undefined \def \showISBNxiii  #1{\unskip}     \fi
\ifx \showISSN     \undefined \def \showISSN      #1{\unskip}     \fi
\ifx \showLCCN     \undefined \def \showLCCN      #1{\unskip}     \fi
\ifx \shownote     \undefined \def \shownote      #1{#1}          \fi
\ifx \showarticletitle \undefined \def \showarticletitle #1{#1}   \fi
\ifx \showURL      \undefined \def \showURL       {\relax}        \fi
\providecommand\bibfield[2]{#2}
\providecommand\bibinfo[2]{#2}
\providecommand\natexlab[1]{#1}
\providecommand\showeprint[2][]{arXiv:#2}

\bibitem[AirTable(2022)]%
        {airtable}
\bibfield{author}{\bibinfo{person}{AirTable}.} \bibinfo{year}{2022}\natexlab{}.
\newblock \bibinfo{title}{Airtable | Everyone's app platform}.
\newblock
\newblock
\newblock
\shownote{\url{https://airtable.com/}}.


\bibitem[Alili et~al\mbox{.}(2009)]%
        {alili2009task}
\bibfield{author}{\bibinfo{person}{Samir Alili}, \bibinfo{person}{Rachid
  Alami}, {and} \bibinfo{person}{Vincent Montreuil}.}
  \bibinfo{year}{2009}\natexlab{}.
\newblock \showarticletitle{A Task Planner for an Autonomous Social Robot}. In
  \bibinfo{booktitle}{\emph{Distributed Autonomous Robotic Systems 8}},
  \bibfield{editor}{\bibinfo{person}{Hajime Asama}, \bibinfo{person}{Haruhisa
  Kurokawa}, \bibinfo{person}{Jun Ota}, {and} \bibinfo{person}{Kosuke
  Sekiyama}} (Eds.). \bibinfo{publisher}{Springer Berlin Heidelberg},
  \bibinfo{address}{Berlin, Heidelberg}, \bibinfo{pages}{335--344}.
\newblock
\showISBNx{978-3-642-00644-9}
\urldef\tempurl%
\url{https://doi.org/10.1007/978-3-642-00644-9_30}
\showDOI{\tempurl}


\bibitem[Alterovitz et~al\mbox{.}(2016)]%
        {alterovitz2016robot}
\bibfield{author}{\bibinfo{person}{Ron Alterovitz}, \bibinfo{person}{Sven
  Koenig}, {and} \bibinfo{person}{Maxim Likhachev}.}
  \bibinfo{year}{2016}\natexlab{}.
\newblock \showarticletitle{Robot Planning in the Real World: Research
  Challenges and Opportunities}.
\newblock \bibinfo{journal}{\emph{AI Magazine}} \bibinfo{volume}{37},
  \bibinfo{number}{2} (\bibinfo{date}{Jul.} \bibinfo{year}{2016}),
  \bibinfo{pages}{76--84}.
\newblock
\urldef\tempurl%
\url{https://doi.org/10.1609/aimag.v37i2.2651}
\showDOI{\tempurl}


\bibitem[Amershi et~al\mbox{.}(2019)]%
        {amershi2019guidelines}
\bibfield{author}{\bibinfo{person}{Saleema Amershi}, \bibinfo{person}{Dan
  Weld}, \bibinfo{person}{Mihaela Vorvoreanu}, \bibinfo{person}{Adam Fourney},
  \bibinfo{person}{Besmira Nushi}, \bibinfo{person}{Penny Collisson},
  \bibinfo{person}{Jina Suh}, \bibinfo{person}{Shamsi Iqbal},
  \bibinfo{person}{Paul~N. Bennett}, \bibinfo{person}{Kori Inkpen},
  \bibinfo{person}{Jaime Teevan}, \bibinfo{person}{Ruth Kikin-Gil}, {and}
  \bibinfo{person}{Eric Horvitz}.} \bibinfo{year}{2019}\natexlab{}.
\newblock \showarticletitle{Guidelines for Human-AI Interaction}. In
  \bibinfo{booktitle}{\emph{Proceedings of the 2019 CHI Conference on Human
  Factors in Computing Systems}} (Glasgow, Scotland Uk)
  \emph{(\bibinfo{series}{CHI '19})}. \bibinfo{publisher}{Association for
  Computing Machinery}, \bibinfo{address}{New York, NY, USA},
  \bibinfo{pages}{1–13}.
\newblock
\showISBNx{9781450359702}
\urldef\tempurl%
\url{https://doi.org/10.1145/3290605.3300233}
\showDOI{\tempurl}


\bibitem[Boniardi et~al\mbox{.}(2016)]%
        {boniardiSketching}
\bibfield{author}{\bibinfo{person}{Federico Boniardi}, \bibinfo{person}{Abhinav
  Valada}, \bibinfo{person}{Wolfram Burgard}, {and} \bibinfo{person}{Gian~Diego
  Tipaldi}.} \bibinfo{year}{2016}\natexlab{}.
\newblock \showarticletitle{Autonomous indoor robot navigation using a sketch
  interface for drawing maps and routes}. In \bibinfo{booktitle}{\emph{2016
  IEEE International Conference on Robotics and Automation (ICRA)}}.
  \bibinfo{pages}{2896--2901}.
\newblock
\urldef\tempurl%
\url{https://doi.org/10.1109/ICRA.2016.7487453}
\showDOI{\tempurl}


\bibitem[Brooke et~al\mbox{.}(1996)]%
        {brooke1996sus}
\bibfield{author}{\bibinfo{person}{John Brooke} {et~al\mbox{.}}}
  \bibinfo{year}{1996}\natexlab{}.
\newblock \showarticletitle{SUS-A quick and dirty usability scale}.
\newblock \bibinfo{journal}{\emph{Usability evaluation in industry}}
  \bibinfo{volume}{189}, \bibinfo{number}{194} (\bibinfo{year}{1996}),
  \bibinfo{pages}{4--7}.
\newblock


\bibitem[Burnett et~al\mbox{.}(2004)]%
        {burnett2004end}
\bibfield{author}{\bibinfo{person}{Margaret Burnett}, \bibinfo{person}{Curtis
  Cook}, {and} \bibinfo{person}{Gregg Rothermel}.}
  \bibinfo{year}{2004}\natexlab{}.
\newblock \showarticletitle{End-User Software Engineering}.
\newblock \bibinfo{journal}{\emph{Commun. ACM}} \bibinfo{volume}{47},
  \bibinfo{number}{9} (\bibinfo{date}{sep} \bibinfo{year}{2004}),
  \bibinfo{pages}{53–58}.
\newblock
\showISSN{0001-0782}
\urldef\tempurl%
\url{https://doi.org/10.1145/1015864.1015889}
\showDOI{\tempurl}


\bibitem[Cao et~al\mbox{.}(2019)]%
        {vra}
\bibfield{author}{\bibinfo{person}{Yuanzhi Cao}, \bibinfo{person}{Zhuangying
  Xu}, \bibinfo{person}{Fan Li}, \bibinfo{person}{Wentao Zhong},
  \bibinfo{person}{Ke Huo}, {and} \bibinfo{person}{Karthik Ramani}.}
  \bibinfo{year}{2019}\natexlab{}.
\newblock \showarticletitle{V.Ra: An In-Situ Visual Authoring System for
  Robot-IoT Task Planning with Augmented Reality}. In
  \bibinfo{booktitle}{\emph{Proceedings of the 2019 on Designing Interactive
  Systems Conference}} (San Diego, CA, USA) \emph{(\bibinfo{series}{DIS '19})}.
  \bibinfo{publisher}{Association for Computing Machinery},
  \bibinfo{address}{New York, NY, USA}, \bibinfo{pages}{1059–1070}.
\newblock
\showISBNx{9781450358507}
\urldef\tempurl%
\url{https://doi.org/10.1145/3322276.3322278}
\showDOI{\tempurl}


\bibitem[Chen and Mooney(2011)]%
        {chen11learning}
\bibfield{author}{\bibinfo{person}{David Chen} {and} \bibinfo{person}{Raymond
  Mooney}.} \bibinfo{year}{2011}\natexlab{}.
\newblock \showarticletitle{Learning to Interpret Natural Language Navigation
  Instructions from Observations}.
\newblock \bibinfo{journal}{\emph{Proceedings of the AAAI Conference on
  Artificial Intelligence}} \bibinfo{volume}{25}, \bibinfo{number}{1}
  (\bibinfo{date}{Aug.} \bibinfo{year}{2011}), \bibinfo{pages}{859--865}.
\newblock
\urldef\tempurl%
\url{https://doi.org/10.1609/aaai.v25i1.7974}
\showDOI{\tempurl}


\bibitem[Chung and Cakmak(2022)]%
        {chung2022authoring}
\bibfield{author}{\bibinfo{person}{Michael Jae-Yoon Chung} {and}
  \bibinfo{person}{Maya Cakmak}.} \bibinfo{year}{2022}\natexlab{}.
\newblock \showarticletitle{Authoring Human Simulators via Probabilistic
  Functional Reactive Program Synthesis}. In \bibinfo{booktitle}{\emph{2022
  17th ACM/IEEE International Conference on Human-Robot Interaction (HRI)}}.
  \bibinfo{pages}{727--730}.
\newblock
\urldef\tempurl%
\url{https://doi.org/10.1109/HRI53351.2022.9889630}
\showDOI{\tempurl}


\bibitem[Connell(2019)]%
        {connell2019verbal}
\bibfield{author}{\bibinfo{person}{Jonathan Connell}.}
  \bibinfo{year}{2019}\natexlab{}.
\newblock \showarticletitle{Verbal Programming of Robot Behavior}.
\newblock \bibinfo{journal}{\emph{arXiv preprint arXiv:1911.09782}}
  (\bibinfo{year}{2019}).
\newblock


\bibitem[Correa et~al\mbox{.}(2010)]%
        {rtcontrol1}
\bibfield{author}{\bibinfo{person}{Andrew Correa}, \bibinfo{person}{Matthew~R.
  Walter}, \bibinfo{person}{Luke Fletcher}, \bibinfo{person}{Jim Glass},
  \bibinfo{person}{Seth Teller}, {and} \bibinfo{person}{Randall Davis}.}
  \bibinfo{year}{2010}\natexlab{}.
\newblock \showarticletitle{Multimodal interaction with an autonomous
  forklift}. In \bibinfo{booktitle}{\emph{2010 5th ACM/IEEE International
  Conference on Human-Robot Interaction (HRI)}}. \bibinfo{pages}{243--250}.
\newblock
\urldef\tempurl%
\url{https://doi.org/10.1109/HRI.2010.5453188}
\showDOI{\tempurl}


\bibitem[Dong and Lapata(2016)]%
        {dong2016language}
\bibfield{author}{\bibinfo{person}{Li Dong} {and} \bibinfo{person}{Mirella
  Lapata}.} \bibinfo{year}{2016}\natexlab{}.
\newblock \showarticletitle{Language to Logical Form with Neural Attention}. In
  \bibinfo{booktitle}{\emph{54th Annual Meeting of the Association for
  Computational Linguistics, ACL 2016 - Long Papers}},
  Vol.~\bibinfo{volume}{1}. \bibinfo{publisher}{Association for Computational
  Linguistics (ACL)}, \bibinfo{pages}{33--43}.
\newblock
\urldef\tempurl%
\url{https://doi.org/10.18653/v1/P16-1004}
\showDOI{\tempurl}
\newblock
\shownote{54th Annual Meeting of the Association for Computational Linguistics,
  ACL 2016 ; Conference date: 07-08-2016 Through 12-08-2016}.


\bibitem[Fellbaum(1998)]%
        {wordnet1998}
\bibfield{author}{\bibinfo{person}{Christiane Fellbaum}.}
  \bibinfo{year}{1998}\natexlab{}.
\newblock \bibinfo{booktitle}{\emph{{WordNet: An Electronic Lexical
  Database}}}.
\newblock \bibinfo{publisher}{The MIT Press}.
\newblock
\showISBNx{9780262272551}
\urldef\tempurl%
\url{https://doi.org/10.7551/mitpress/7287.001.0001}
\showDOI{\tempurl}


\bibitem[Forbes et~al\mbox{.}(2015)]%
        {forbes2015robot}
\bibfield{author}{\bibinfo{person}{Maxwell Forbes}, \bibinfo{person}{Rajesh
  P.~N. Rao}, \bibinfo{person}{Luke Zettlemoyer}, {and} \bibinfo{person}{Maya
  Cakmak}.} \bibinfo{year}{2015}\natexlab{}.
\newblock \showarticletitle{Robot Programming by Demonstration with situated
  spatial language understanding}. In \bibinfo{booktitle}{\emph{2015 IEEE
  International Conference on Robotics and Automation (ICRA)}}.
  \bibinfo{pages}{2014--2020}.
\newblock
\urldef\tempurl%
\url{https://doi.org/10.1109/ICRA.2015.7139462}
\showDOI{\tempurl}


\bibitem[Gao and Huang(2019)]%
        {gao2019pati}
\bibfield{author}{\bibinfo{person}{Yuxiang Gao} {and}
  \bibinfo{person}{Chien-Ming Huang}.} \bibinfo{year}{2019}\natexlab{}.
\newblock \showarticletitle{PATI: A Projection-Based Augmented Table-Top
  Interface for Robot Programming}. In \bibinfo{booktitle}{\emph{Proceedings of
  the 24th International Conference on Intelligent User Interfaces}} (Marina
  del Ray, California) \emph{(\bibinfo{series}{IUI '19})}.
  \bibinfo{publisher}{Association for Computing Machinery},
  \bibinfo{address}{New York, NY, USA}, \bibinfo{pages}{345–355}.
\newblock
\showISBNx{9781450362726}
\urldef\tempurl%
\url{https://doi.org/10.1145/3301275.3302326}
\showDOI{\tempurl}


\bibitem[Ghallab et~al\mbox{.}(2016)]%
        {ghallab2016automated}
\bibfield{author}{\bibinfo{person}{Malik Ghallab}, \bibinfo{person}{Dana Nau},
  {and} \bibinfo{person}{Paolo Traverso}.} \bibinfo{year}{2016}\natexlab{}.
\newblock \bibinfo{booktitle}{\emph{Automated Planning and Acting}}.
\newblock \bibinfo{publisher}{Cambridge University Press}.
\newblock
\urldef\tempurl%
\url{https://doi.org/10.1017/CBO9781139583923}
\showDOI{\tempurl}


\bibitem[Gorostiza and Salichs(2011)]%
        {gorostiza2011end}
\bibfield{author}{\bibinfo{person}{Javi~F. Gorostiza} {and}
  \bibinfo{person}{Miguel~A. Salichs}.} \bibinfo{year}{2011}\natexlab{}.
\newblock \showarticletitle{End-User Programming of a Social Robot by Dialog}.
\newblock \bibinfo{journal}{\emph{Robot. Auton. Syst.}} \bibinfo{volume}{59},
  \bibinfo{number}{12} (\bibinfo{date}{dec} \bibinfo{year}{2011}),
  \bibinfo{pages}{1102–1114}.
\newblock
\showISSN{0921-8890}
\urldef\tempurl%
\url{https://doi.org/10.1016/j.robot.2011.07.009}
\showDOI{\tempurl}


\bibitem[Gulwani et~al\mbox{.}(2017)]%
        {gulwani2017program}
\bibfield{author}{\bibinfo{person}{Sumit Gulwani}, \bibinfo{person}{Oleksandr
  Polozov}, {and} \bibinfo{person}{Rishabh Singh}.}
  \bibinfo{year}{2017}\natexlab{}.
\newblock \showarticletitle{Program Synthesis}.
\newblock \bibinfo{journal}{\emph{Foundations and Trends® in Programming
  Languages}} \bibinfo{volume}{4}, \bibinfo{number}{1-2}
  (\bibinfo{year}{2017}), \bibinfo{pages}{1--119}.
\newblock
\showISSN{2325-1107}
\urldef\tempurl%
\url{https://doi.org/10.1561/2500000010}
\showDOI{\tempurl}


\bibitem[Huang et~al\mbox{.}(2019)]%
        {huang2019synthesizing}
\bibfield{author}{\bibinfo{person}{Justin Huang}, \bibinfo{person}{Dieter Fox},
  {and} \bibinfo{person}{Maya Cakmak}.} \bibinfo{year}{2019}\natexlab{}.
\newblock \showarticletitle{Synthesizing Robot Manipulation Programs from a
  Single Observed Human Demonstration}. In \bibinfo{booktitle}{\emph{2019
  IEEE/RSJ International Conference on Intelligent Robots and Systems (IROS)}}.
  \bibinfo{pages}{4585--4592}.
\newblock
\urldef\tempurl%
\url{https://doi.org/10.1109/IROS40897.2019.8968543}
\showDOI{\tempurl}


\bibitem[Huang et~al\mbox{.}(2016)]%
        {huang2016design}
\bibfield{author}{\bibinfo{person}{Justin Huang}, \bibinfo{person}{Tessa Lau},
  {and} \bibinfo{person}{Maya Cakmak}.} \bibinfo{year}{2016}\natexlab{}.
\newblock \showarticletitle{Design and evaluation of a rapid programming system
  for service robots}. In \bibinfo{booktitle}{\emph{2016 11th ACM/IEEE
  International Conference on Human-Robot Interaction (HRI)}}.
  \bibinfo{pages}{295--302}.
\newblock
\urldef\tempurl%
\url{https://doi.org/10.1109/HRI.2016.7451765}
\showDOI{\tempurl}


\bibitem[Kipper et~al\mbox{.}(2000)]%
        {kipper2000class}
\bibfield{author}{\bibinfo{person}{Karin Kipper}, \bibinfo{person}{Hoa~Trang
  Dang}, \bibinfo{person}{Martha Palmer}, {et~al\mbox{.}}}
  \bibinfo{year}{2000}\natexlab{}.
\newblock \showarticletitle{Class-based construction of a verb lexicon}.
\newblock \bibinfo{journal}{\emph{AAAI/IAAI}}  \bibinfo{volume}{691}
  (\bibinfo{year}{2000}), \bibinfo{pages}{696}.
\newblock


\bibitem[Kirk and Laird(2019)]%
        {kirk2019learning}
\bibfield{author}{\bibinfo{person}{James~R. Kirk} {and}
  \bibinfo{person}{John~E. Laird}.} \bibinfo{year}{2019}\natexlab{}.
\newblock \showarticletitle{Learning Hierarchical Symbolic Representations to
  Support Interactive Task Learning and Knowledge Transfer}. In
  \bibinfo{booktitle}{\emph{Proceedings of the Twenty-Eighth International
  Joint Conference on Artificial Intelligence, {IJCAI-19}}}.
  \bibinfo{publisher}{International Joint Conferences on Artificial
  Intelligence Organization}, \bibinfo{pages}{6095--6102}.
\newblock
\urldef\tempurl%
\url{https://doi.org/10.24963/ijcai.2019/844}
\showDOI{\tempurl}


\bibitem[Kress-Gazit et~al\mbox{.}(2008)]%
        {kress2008translating}
\bibfield{author}{\bibinfo{person}{Hadas Kress-Gazit},
  \bibinfo{person}{Georgios~E. Fainekos}, {and} \bibinfo{person}{George~J.
  Pappas}.} \bibinfo{year}{2008}\natexlab{}.
\newblock \showarticletitle{Translating Structured English to Robot
  Controllers}.
\newblock \bibinfo{journal}{\emph{Advanced Robotics}} \bibinfo{volume}{22},
  \bibinfo{number}{12} (\bibinfo{year}{2008}), \bibinfo{pages}{1343--1359}.
\newblock
\urldef\tempurl%
\url{https://doi.org/10.1163/156855308X344864}
\showDOI{\tempurl}
\showeprint{https://doi.org/10.1163/156855308X344864}


\bibitem[Lauria et~al\mbox{.}(2002)]%
        {lauria2002mobile}
\bibfield{author}{\bibinfo{person}{Stanislao Lauria}, \bibinfo{person}{Guido
  Bugmann}, \bibinfo{person}{Theocharis Kyriacou}, {and} \bibinfo{person}{Ewan
  Klein}.} \bibinfo{year}{2002}\natexlab{}.
\newblock \showarticletitle{Mobile robot programming using natural language}.
\newblock \bibinfo{journal}{\emph{Robotics and Autonomous Systems}}
  \bibinfo{volume}{38}, \bibinfo{number}{3} (\bibinfo{year}{2002}),
  \bibinfo{pages}{171--181}.
\newblock
\showISSN{0921-8890}
\urldef\tempurl%
\url{https://doi.org/10.1016/S0921-8890(02)00166-5}
\showDOI{\tempurl}
\newblock
\shownote{Advances in Robot Skill Learning}.


\bibitem[Leonardi et~al\mbox{.}(2019)]%
        {leonardi2019trigger}
\bibfield{author}{\bibinfo{person}{Nicola Leonardi}, \bibinfo{person}{Marco
  Manca}, \bibinfo{person}{Fabio Patern\`{o}}, {and} \bibinfo{person}{Carmen
  Santoro}.} \bibinfo{year}{2019}\natexlab{}.
\newblock \showarticletitle{Trigger-Action Programming for Personalising
  Humanoid Robot Behaviour}. In \bibinfo{booktitle}{\emph{Proceedings of the
  2019 CHI Conference on Human Factors in Computing Systems}} (Glasgow,
  Scotland Uk) \emph{(\bibinfo{series}{CHI '19})}.
  \bibinfo{publisher}{Association for Computing Machinery},
  \bibinfo{address}{New York, NY, USA}, \bibinfo{pages}{1–13}.
\newblock
\showISBNx{9781450359702}
\urldef\tempurl%
\url{https://doi.org/10.1145/3290605.3300675}
\showDOI{\tempurl}


\bibitem[Lieberman et~al\mbox{.}(2006)]%
        {lieberman2006end}
\bibfield{author}{\bibinfo{person}{Henry Lieberman}, \bibinfo{person}{Fabio
  Patern{\`o}}, \bibinfo{person}{Markus Klann}, {and} \bibinfo{person}{Volker
  Wulf}.} \bibinfo{year}{2006}\natexlab{}.
\newblock \bibinfo{booktitle}{\emph{End-User Development: An Emerging
  Paradigm}}.
\newblock \bibinfo{publisher}{Springer Netherlands},
  \bibinfo{address}{Dordrecht}, \bibinfo{pages}{1--8}.
\newblock
\showISBNx{978-1-4020-5386-3}
\urldef\tempurl%
\url{https://doi.org/10.1007/1-4020-5386-X_1}
\showDOI{\tempurl}


\bibitem[Little et~al\mbox{.}(2010)]%
        {little2011sloppy}
\bibfield{author}{\bibinfo{person}{Greg Little}, \bibinfo{person}{Robert~C.
  Miller}, \bibinfo{person}{Victoria~H. Chou}, \bibinfo{person}{Michael
  Bernstein}, \bibinfo{person}{Tessa Lau}, {and} \bibinfo{person}{Allen
  Cypher}.} \bibinfo{year}{2010}\natexlab{}.
\newblock \showarticletitle{Sloppy Programming}.
\newblock In \bibinfo{booktitle}{\emph{No Code Required}},
  \bibfield{editor}{\bibinfo{person}{Allen Cypher}, \bibinfo{person}{Mira
  Dontcheva}, \bibinfo{person}{Tessa Lau}, {and} \bibinfo{person}{Jeffrey
  Nichols}} (Eds.). \bibinfo{publisher}{Morgan Kaufmann},
  \bibinfo{address}{Boston}, \bibinfo{pages}{289--307}.
\newblock
\showISBNx{978-0-12-381541-5}
\urldef\tempurl%
\url{https://doi.org/10.1016/B978-0-12-381541-5.00015-8}
\showDOI{\tempurl}


\bibitem[Liu et~al\mbox{.}(2011)]%
        {liu2011roboshop}
\bibfield{author}{\bibinfo{person}{Kexi Liu}, \bibinfo{person}{Daisuke
  Sakamoto}, \bibinfo{person}{Masahiko Inami}, {and} \bibinfo{person}{Takeo
  Igarashi}.} \bibinfo{year}{2011}\natexlab{}.
\newblock \showarticletitle{Roboshop: Multi-Layered Sketching Interface for
  Robot Housework Assignment and Management}. In
  \bibinfo{booktitle}{\emph{Proceedings of the SIGCHI Conference on Human
  Factors in Computing Systems}} (Vancouver, BC, Canada)
  \emph{(\bibinfo{series}{CHI '11})}. \bibinfo{publisher}{Association for
  Computing Machinery}, \bibinfo{address}{New York, NY, USA},
  \bibinfo{pages}{647–656}.
\newblock
\showISBNx{9781450302289}
\urldef\tempurl%
\url{https://doi.org/10.1145/1978942.1979035}
\showDOI{\tempurl}


\bibitem[Lund(2001)]%
        {lund2001measuring}
\bibfield{author}{\bibinfo{person}{Arnold~M Lund}.}
  \bibinfo{year}{2001}\natexlab{}.
\newblock \showarticletitle{Measuring Usability with the USE Questionnaire}.
\newblock \bibinfo{journal}{\emph{Usability and User Experience Newsletter of
  the STC Usability SIG}} \bibinfo{volume}{8}, \bibinfo{number}{2}
  (\bibinfo{date}{01} \bibinfo{year}{2001}), \bibinfo{pages}{3--6}.
\newblock


\bibitem[Manning et~al\mbox{.}(2014)]%
        {manning2014stanford}
\bibfield{author}{\bibinfo{person}{Christopher~D. Manning},
  \bibinfo{person}{Mihai Surdeanu}, \bibinfo{person}{John Bauer},
  \bibinfo{person}{Jenny Finkel}, \bibinfo{person}{Steven~J. Bethard}, {and}
  \bibinfo{person}{David McClosky}.} \bibinfo{year}{2014}\natexlab{}.
\newblock \showarticletitle{The {Stanford} {CoreNLP} Natural Language
  Processing Toolkit}. In \bibinfo{booktitle}{\emph{Association for
  Computational Linguistics (ACL) System Demonstrations}}.
  \bibinfo{pages}{55--60}.
\newblock
\urldef\tempurl%
\url{http://www.aclweb.org/anthology/P/P14/P14-5010}
\showURL{%
\tempurl}


\bibitem[Matuszek et~al\mbox{.}(2013)]%
        {matuszek2013}
\bibfield{author}{\bibinfo{person}{Cynthia Matuszek}, \bibinfo{person}{Evan
  Herbst}, \bibinfo{person}{Luke Zettlemoyer}, {and} \bibinfo{person}{Dieter
  Fox}.} \bibinfo{year}{2013}\natexlab{}.
\newblock \bibinfo{booktitle}{\emph{Learning to Parse Natural Language Commands
  to a Robot Control System}}.
\newblock \bibinfo{publisher}{Springer International Publishing},
  \bibinfo{address}{Heidelberg}, \bibinfo{pages}{403--415}.
\newblock
\showISBNx{978-3-319-00065-7}
\urldef\tempurl%
\url{https://doi.org/10.1007/978-3-319-00065-7_28}
\showDOI{\tempurl}


\bibitem[Miller(1995)]%
        {wordnet1995}
\bibfield{author}{\bibinfo{person}{George~A. Miller}.}
  \bibinfo{year}{1995}\natexlab{}.
\newblock \showarticletitle{WordNet: A Lexical Database for English}.
\newblock \bibinfo{journal}{\emph{Commun. ACM}} \bibinfo{volume}{38},
  \bibinfo{number}{11} (\bibinfo{date}{nov} \bibinfo{year}{1995}),
  \bibinfo{pages}{39–41}.
\newblock
\showISSN{0001-0782}
\urldef\tempurl%
\url{https://doi.org/10.1145/219717.219748}
\showDOI{\tempurl}


\bibitem[Mohan and Laird(2014)]%
        {mohan2014learning}
\bibfield{author}{\bibinfo{person}{Shiwali Mohan} {and} \bibinfo{person}{John
  Laird}.} \bibinfo{year}{2014}\natexlab{}.
\newblock \showarticletitle{Learning Goal-Oriented Hierarchical Tasks from
  Situated Interactive Instruction}.
\newblock \bibinfo{journal}{\emph{Proceedings of the AAAI Conference on
  Artificial Intelligence}} \bibinfo{volume}{28}, \bibinfo{number}{1}
  (\bibinfo{date}{Jun.} \bibinfo{year}{2014}).
\newblock
\urldef\tempurl%
\url{https://doi.org/10.1609/aaai.v28i1.8756}
\showDOI{\tempurl}


\bibitem[Petrick and Foster(2013)]%
        {petrick2013planning}
\bibfield{author}{\bibinfo{person}{Ronald Petrick} {and}
  \bibinfo{person}{Mary~Ellen Foster}.} \bibinfo{year}{2013}\natexlab{}.
\newblock \showarticletitle{Planning for Social Interaction in a Robot
  Bartender Domain}.
\newblock \bibinfo{journal}{\emph{Proceedings of the International Conference
  on Automated Planning and Scheduling}} \bibinfo{volume}{23},
  \bibinfo{number}{1} (\bibinfo{date}{Jun.} \bibinfo{year}{2013}),
  \bibinfo{pages}{389--397}.
\newblock
\urldef\tempurl%
\url{https://doi.org/10.1609/icaps.v23i1.13589}
\showDOI{\tempurl}


\bibitem[Porfirio et~al\mbox{.}(2018)]%
        {porfirio2018authoring}
\bibfield{author}{\bibinfo{person}{David Porfirio}, \bibinfo{person}{Allison
  Saupp\'{e}}, \bibinfo{person}{Aws Albarghouthi}, {and} \bibinfo{person}{Bilge
  Mutlu}.} \bibinfo{year}{2018}\natexlab{}.
\newblock \showarticletitle{Authoring and Verifying Human-Robot Interactions}.
  In \bibinfo{booktitle}{\emph{Proceedings of the 31st Annual ACM Symposium on
  User Interface Software and Technology}} (Berlin, Germany)
  \emph{(\bibinfo{series}{UIST '18})}. \bibinfo{publisher}{Association for
  Computing Machinery}, \bibinfo{address}{New York, NY, USA},
  \bibinfo{pages}{75–86}.
\newblock
\showISBNx{9781450359481}
\urldef\tempurl%
\url{https://doi.org/10.1145/3242587.3242634}
\showDOI{\tempurl}


\bibitem[Porfirio(2022)]%
        {porfirio2022authoring}
\bibfield{author}{\bibinfo{person}{David~J Porfirio}.}
  \bibinfo{year}{2022}\natexlab{}.
\newblock \emph{\bibinfo{title}{Authoring Social Interactions Between Humans
  and Robots}}.
\newblock \bibinfo{thesistype}{Ph.\,D. Dissertation}.
  \bibinfo{school}{UW--Madison}.
\newblock


\bibitem[Porfirio et~al\mbox{.}(2021)]%
        {porfirio2021figaro}
\bibfield{author}{\bibinfo{person}{David~J. Porfirio}, \bibinfo{person}{Laura
  Stegner}, \bibinfo{person}{Maya Cakmak}, \bibinfo{person}{Allison
  Saupp\'{e}}, \bibinfo{person}{Aws Albarghouthi}, {and} \bibinfo{person}{Bilge
  Mutlu}.} \bibinfo{year}{2021}\natexlab{}.
\newblock \showarticletitle{Figaro: A Tabletop Authoring Environment for
  Human-Robot Interaction}. In \bibinfo{booktitle}{\emph{Proceedings of the
  2021 CHI Conference on Human Factors in Computing Systems}} (Yokohama, Japan)
  \emph{(\bibinfo{series}{CHI '21})}. \bibinfo{publisher}{Association for
  Computing Machinery}, \bibinfo{address}{New York, NY, USA}, Article
  \bibinfo{articleno}{414}, \bibinfo{numpages}{15}~pages.
\newblock
\showISBNx{9781450380966}
\urldef\tempurl%
\url{https://doi.org/10.1145/3411764.3446864}
\showDOI{\tempurl}


\bibitem[Resnick et~al\mbox{.}(2009)]%
        {resnick2009scratch}
\bibfield{author}{\bibinfo{person}{Mitchel Resnick}, \bibinfo{person}{John
  Maloney}, \bibinfo{person}{Andr\'{e}s Monroy-Hern\'{a}ndez},
  \bibinfo{person}{Natalie Rusk}, \bibinfo{person}{Evelyn Eastmond},
  \bibinfo{person}{Karen Brennan}, \bibinfo{person}{Amon Millner},
  \bibinfo{person}{Eric Rosenbaum}, \bibinfo{person}{Jay Silver},
  \bibinfo{person}{Brian Silverman}, {and} \bibinfo{person}{Yasmin Kafai}.}
  \bibinfo{year}{2009}\natexlab{}.
\newblock \showarticletitle{Scratch: Programming for All}.
\newblock \bibinfo{journal}{\emph{Commun. ACM}} \bibinfo{volume}{52},
  \bibinfo{number}{11} (\bibinfo{date}{nov} \bibinfo{year}{2009}),
  \bibinfo{pages}{60–67}.
\newblock
\showISSN{0001-0782}
\urldef\tempurl%
\url{https://doi.org/10.1145/1592761.1592779}
\showDOI{\tempurl}


\bibitem[Sakamoto et~al\mbox{.}(2009)]%
        {sakamoto2009sketch}
\bibfield{author}{\bibinfo{person}{Daisuke Sakamoto}, \bibinfo{person}{Koichiro
  Honda}, \bibinfo{person}{Masahiko Inami}, {and} \bibinfo{person}{Takeo
  Igarashi}.} \bibinfo{year}{2009}\natexlab{}.
\newblock \showarticletitle{Sketch and Run: A Stroke-Based Interface for Home
  Robots}. In \bibinfo{booktitle}{\emph{Proceedings of the SIGCHI Conference on
  Human Factors in Computing Systems}} (Boston, MA, USA)
  \emph{(\bibinfo{series}{CHI '09})}. \bibinfo{publisher}{Association for
  Computing Machinery}, \bibinfo{address}{New York, NY, USA},
  \bibinfo{pages}{197–200}.
\newblock
\showISBNx{9781605582467}
\urldef\tempurl%
\url{https://doi.org/10.1145/1518701.1518733}
\showDOI{\tempurl}


\bibitem[Senft et~al\mbox{.}(2021)]%
        {senft2021situated}
\bibfield{author}{\bibinfo{person}{Emmanuel Senft}, \bibinfo{person}{Michael
  Hagenow}, \bibinfo{person}{Robert Radwin}, \bibinfo{person}{Michael Zinn},
  \bibinfo{person}{Michael Gleicher}, {and} \bibinfo{person}{Bilge Mutlu}.}
  \bibinfo{year}{2021}\natexlab{}.
\newblock \showarticletitle{Situated Live Programming for Human-Robot
  Collaboration}. In \bibinfo{booktitle}{\emph{The 34th Annual ACM Symposium on
  User Interface Software and Technology}} (Virtual Event, USA)
  \emph{(\bibinfo{series}{UIST '21})}. \bibinfo{publisher}{Association for
  Computing Machinery}, \bibinfo{address}{New York, NY, USA},
  \bibinfo{pages}{613–625}.
\newblock
\showISBNx{9781450386357}
\urldef\tempurl%
\url{https://doi.org/10.1145/3472749.3474773}
\showDOI{\tempurl}


\bibitem[Shah(2012)]%
        {shah2012towards}
\bibfield{author}{\bibinfo{person}{Danelle Shah}.}
  \bibinfo{year}{2012}\natexlab{}.
\newblock \emph{\bibinfo{title}{Towards Natural And Robust Human-Robot
  Interaction Using Sketch And Speech}}.
\newblock \bibinfo{thesistype}{Ph.\,D. Dissertation}.
  \bibinfo{school}{Cornell}.
\newblock


\bibitem[Shah et~al\mbox{.}(2010)]%
        {shah2010}
\bibfield{author}{\bibinfo{person}{Danelle Shah}, \bibinfo{person}{Joseph
  Schneider}, {and} \bibinfo{person}{Mark Campbell}.}
  \bibinfo{year}{2010}\natexlab{}.
\newblock \showarticletitle{A robust sketch interface for natural robot
  control}. In \bibinfo{booktitle}{\emph{2010 IEEE/RSJ International Conference
  on Intelligent Robots and Systems}}. \bibinfo{pages}{4458--4463}.
\newblock
\urldef\tempurl%
\url{https://doi.org/10.1109/IROS.2010.5649345}
\showDOI{\tempurl}


\bibitem[Stenmark and Nugues(2013)]%
        {stenmark2013natural}
\bibfield{author}{\bibinfo{person}{Maj Stenmark} {and} \bibinfo{person}{Pierre
  Nugues}.} \bibinfo{year}{2013}\natexlab{}.
\newblock \showarticletitle{Natural language programming of industrial robots}.
  In \bibinfo{booktitle}{\emph{IEEE ISR 2013}}. \bibinfo{pages}{1--5}.
\newblock
\urldef\tempurl%
\url{https://doi.org/10.1109/ISR.2013.6695630}
\showDOI{\tempurl}


\bibitem[Teller et~al\mbox{.}(2010)]%
        {rtcontrol2}
\bibfield{author}{\bibinfo{person}{Seth Teller}, \bibinfo{person}{Matthew~R.
  Walter}, \bibinfo{person}{Matthew Antone}, \bibinfo{person}{Andrew Correa},
  \bibinfo{person}{Randall Davis}, \bibinfo{person}{Luke Fletcher},
  \bibinfo{person}{Emilio Frazzoli}, \bibinfo{person}{Jim Glass},
  \bibinfo{person}{Jonathan~P. How}, \bibinfo{person}{Albert~S. Huang},
  \bibinfo{person}{Jeong~hwan Jeon}, \bibinfo{person}{Sertac Karaman},
  \bibinfo{person}{Brandon Luders}, \bibinfo{person}{Nicholas Roy}, {and}
  \bibinfo{person}{Tara Sainath}.} \bibinfo{year}{2010}\natexlab{}.
\newblock \showarticletitle{A voice-commandable robotic forklift working
  alongside humans in minimally-prepared outdoor environments}. In
  \bibinfo{booktitle}{\emph{2010 IEEE International Conference on Robotics and
  Automation}}. \bibinfo{pages}{526--533}.
\newblock
\urldef\tempurl%
\url{https://doi.org/10.1109/ROBOT.2010.5509238}
\showDOI{\tempurl}


\bibitem[Tellex et~al\mbox{.}(2011)]%
        {tellex2011understanding}
\bibfield{author}{\bibinfo{person}{Stefanie Tellex}, \bibinfo{person}{Thomas
  Kollar}, \bibinfo{person}{Steven Dickerson}, \bibinfo{person}{Matthew
  Walter}, \bibinfo{person}{Ashis Banerjee}, \bibinfo{person}{Seth Teller},
  {and} \bibinfo{person}{Nicholas Roy}.} \bibinfo{year}{2011}\natexlab{}.
\newblock \showarticletitle{Understanding Natural Language Commands for Robotic
  Navigation and Mobile Manipulation}.
\newblock \bibinfo{journal}{\emph{Proceedings of the AAAI Conference on
  Artificial Intelligence}} \bibinfo{volume}{25}, \bibinfo{number}{1}
  (\bibinfo{date}{Aug.} \bibinfo{year}{2011}), \bibinfo{pages}{1507--1514}.
\newblock
\urldef\tempurl%
\url{https://doi.org/10.1609/aaai.v25i1.7979}
\showDOI{\tempurl}


\bibitem[Thomason et~al\mbox{.}(2019)]%
        {thomason2019improving}
\bibfield{author}{\bibinfo{person}{Jesse Thomason}, \bibinfo{person}{Aishwarya
  Padmakumar}, \bibinfo{person}{Jivko Sinapov}, \bibinfo{person}{Nick Walker},
  \bibinfo{person}{Yuqian Jiang}, \bibinfo{person}{Harel Yedidsion},
  \bibinfo{person}{Justin Hart}, \bibinfo{person}{Peter Stone}, {and}
  \bibinfo{person}{Raymond~J. Mooney}.} \bibinfo{year}{2019}\natexlab{}.
\newblock \showarticletitle{Improving Grounded Natural Language Understanding
  through Human-Robot Dialog}. In \bibinfo{booktitle}{\emph{2019 International
  Conference on Robotics and Automation (ICRA)}}. \bibinfo{pages}{6934--6941}.
\newblock
\urldef\tempurl%
\url{https://doi.org/10.1109/ICRA.2019.8794287}
\showDOI{\tempurl}


\bibitem[Ur et~al\mbox{.}(2016)]%
        {ur2016trigger}
\bibfield{author}{\bibinfo{person}{Blase Ur}, \bibinfo{person}{Melwyn Pak
  Yong~Ho}, \bibinfo{person}{Stephen Brawner}, \bibinfo{person}{Jiyun Lee},
  \bibinfo{person}{Sarah Mennicken}, \bibinfo{person}{Noah Picard},
  \bibinfo{person}{Diane Schulze}, {and} \bibinfo{person}{Michael~L. Littman}.}
  \bibinfo{year}{2016}\natexlab{}.
\newblock \showarticletitle{Trigger-Action Programming in the Wild: An Analysis
  of 200,000 IFTTT Recipes}. In \bibinfo{booktitle}{\emph{Proceedings of the
  2016 CHI Conference on Human Factors in Computing Systems}} (San Jose,
  California, USA) \emph{(\bibinfo{series}{CHI '16})}.
  \bibinfo{publisher}{Association for Computing Machinery},
  \bibinfo{address}{New York, NY, USA}, \bibinfo{pages}{3227–3231}.
\newblock
\showISBNx{9781450333627}
\urldef\tempurl%
\url{https://doi.org/10.1145/2858036.2858556}
\showDOI{\tempurl}


\bibitem[Walker et~al\mbox{.}(2019)]%
        {walker2019neural}
\bibfield{author}{\bibinfo{person}{Nick Walker}, \bibinfo{person}{Yu-Tang
  Peng}, {and} \bibinfo{person}{Maya Cakmak}.} \bibinfo{year}{2019}\natexlab{}.
\newblock \showarticletitle{Neural Semantic Parsing with Anonymization for
  Command Understanding in General-Purpose Service Robots}. In
  \bibinfo{booktitle}{\emph{RoboCup 2019: Robot World Cup XXIII}},
  \bibfield{editor}{\bibinfo{person}{Stephan Chalup}, \bibinfo{person}{Tim
  Niemueller}, \bibinfo{person}{Jackrit Suthakorn}, {and}
  \bibinfo{person}{Mary-Anne Williams}} (Eds.). \bibinfo{publisher}{Springer
  International Publishing}, \bibinfo{address}{Cham},
  \bibinfo{pages}{337--350}.
\newblock
\showISBNx{978-3-030-35699-6}
\urldef\tempurl%
\url{https://doi.org/10.1007/978-3-030-35699-6_26}
\showDOI{\tempurl}


\bibitem[Webflow(2022)]%
        {webflow}
\bibfield{author}{\bibinfo{person}{Webflow}.} \bibinfo{year}{2022}\natexlab{}.
\newblock \bibinfo{title}{Create a custom website: No-code website builder}.
\newblock
\newblock
\newblock
\shownote{\url{https://webflow.com/}}.


\bibitem[Woods(1973)]%
        {woods1973progress}
\bibfield{author}{\bibinfo{person}{W.~A. Woods}.}
  \bibinfo{year}{1973}\natexlab{}.
\newblock \showarticletitle{Progress in Natural Language Understanding: An
  Application to Lunar Geology}. In \bibinfo{booktitle}{\emph{Proceedings of
  the June 4-8, 1973, National Computer Conference and Exposition}} (New York,
  New York) \emph{(\bibinfo{series}{AFIPS '73})}.
  \bibinfo{publisher}{Association for Computing Machinery},
  \bibinfo{address}{New York, NY, USA}, \bibinfo{pages}{441–450}.
\newblock
\showISBNx{9781450379168}
\urldef\tempurl%
\url{https://doi.org/10.1145/1499586.1499695}
\showDOI{\tempurl}


\bibitem[Worsley et~al\mbox{.}(2018)]%
        {worsley2018multimodal}
\bibfield{author}{\bibinfo{person}{Marcelo Worsley}, \bibinfo{person}{David
  Barel}, \bibinfo{person}{Lydia Davison}, \bibinfo{person}{Thomas Large},
  {and} \bibinfo{person}{Timothy Mwiti}.} \bibinfo{year}{2018}\natexlab{}.
\newblock \showarticletitle{Multimodal Interfaces for Inclusive Learning}. In
  \bibinfo{booktitle}{\emph{Artificial Intelligence in Education}},
  \bibfield{editor}{\bibinfo{person}{Carolyn Penstein~Ros{\'e}},
  \bibinfo{person}{Roberto Mart{\'i}nez-Maldonado}, \bibinfo{person}{H.~Ulrich
  Hoppe}, \bibinfo{person}{Rose Luckin}, \bibinfo{person}{Manolis Mavrikis},
  \bibinfo{person}{Kaska Porayska-Pomsta}, \bibinfo{person}{Bruce McLaren},
  {and} \bibinfo{person}{Benedict du~Boulay}} (Eds.).
  \bibinfo{publisher}{Springer International Publishing},
  \bibinfo{address}{Cham}, \bibinfo{pages}{389--393}.
\newblock
\showISBNx{978-3-319-93846-2}
\urldef\tempurl%
\url{https://doi.org/10.1007/978-3-319-93846-2_73}
\showDOI{\tempurl}


\bibitem[Young et~al\mbox{.}(2012)]%
        {young2012style}
\bibfield{author}{\bibinfo{person}{James Young}, \bibinfo{person}{Kentaro
  Ishii}, \bibinfo{person}{Takeo Igarashi}, {and} \bibinfo{person}{Ehud
  Sharlin}.} \bibinfo{year}{2012}\natexlab{}.
\newblock \showarticletitle{Style by Demonstration: Teaching Interactive
  Movement Style to Robots}. In \bibinfo{booktitle}{\emph{Proceedings of the
  2012 ACM International Conference on Intelligent User Interfaces}} (Lisbon,
  Portugal) \emph{(\bibinfo{series}{IUI '12})}. \bibinfo{publisher}{Association
  for Computing Machinery}, \bibinfo{address}{New York, NY, USA},
  \bibinfo{pages}{41–50}.
\newblock
\showISBNx{9781450310482}
\urldef\tempurl%
\url{https://doi.org/10.1145/2166966.2166976}
\showDOI{\tempurl}


\bibitem[Zapier(2022)]%
        {zapier}
\bibfield{author}{\bibinfo{person}{Zapier}.} \bibinfo{year}{2022}\natexlab{}.
\newblock \bibinfo{title}{Automation that moves you forward}.
\newblock
\newblock
\newblock
\shownote{\url{https://zapier.com/}}.


\bibitem[Zettlemoyer and Collins(2005)]%
        {zettlemoyer2005}
\bibfield{author}{\bibinfo{person}{Luke~S. Zettlemoyer} {and}
  \bibinfo{person}{Michael Collins}.} \bibinfo{year}{2005}\natexlab{}.
\newblock \showarticletitle{Learning to Map Sentences to Logical Form:
  Structured Classification with Probabilistic Categorial Grammars}. In
  \bibinfo{booktitle}{\emph{Proceedings of the Twenty-First Conference on
  Uncertainty in Artificial Intelligence}} (Edinburgh, Scotland)
  \emph{(\bibinfo{series}{UAI'05})}. \bibinfo{publisher}{AUAI Press},
  \bibinfo{address}{Arlington, Virginia, USA}, \bibinfo{pages}{658–666}.
\newblock
\showISBNx{0974903914}


\bibitem[Zhang et~al\mbox{.}(2019)]%
        {zhang2019autotap}
\bibfield{author}{\bibinfo{person}{Lefan Zhang}, \bibinfo{person}{Weijia He},
  \bibinfo{person}{Jesse Martinez}, \bibinfo{person}{Noah Brackenbury},
  \bibinfo{person}{Shan Lu}, {and} \bibinfo{person}{Blase Ur}.}
  \bibinfo{year}{2019}\natexlab{}.
\newblock \showarticletitle{AutoTap: Synthesizing and Repairing Trigger-Action
  Programs Using LTL Properties}. In \bibinfo{booktitle}{\emph{Proceedings of
  the 41st International Conference on Software Engineering}} (Montreal,
  Quebec, Canada) \emph{(\bibinfo{series}{ICSE '19})}. \bibinfo{publisher}{IEEE
  Press}, \bibinfo{pages}{281–291}.
\newblock
\urldef\tempurl%
\url{https://doi.org/10.1109/ICSE.2019.00043}
\showDOI{\tempurl}


\end{thebibliography}

\end{document}